%% file: main.tex
\pdfoutput=1

\documentclass[11pt]{article}

\usepackage[preprint]{acl}

\usepackage{times}
\usepackage{latexsym}

\usepackage[T1]{fontenc}

\usepackage[utf8]{inputenc}

\usepackage{microtype}

\usepackage{inconsolata}

\usepackage{graphicx}

\usepackage{pifont}

\usepackage{enumitem}
\usepackage{booktabs} 
\usepackage{multirow} 
\usepackage{geometry}
\usepackage{graphicx}
\usepackage{algorithm}
\usepackage{algpseudocode}
\usepackage{amsmath}
\usepackage{xcolor}
\usepackage[table]{xcolor}  
\usepackage{pifont}
\usepackage{tcolorbox}  
\usepackage{changepage}
\usepackage{listings}
\usepackage{hyperref}
\usepackage{balance}
\tcbset{
  colframe=black!75,
  colback=gray!10,
  boxrule=0.5pt,
  arc=3pt,
  left=3pt, right=3pt, top=3pt, bottom=3pt,
  fonttitle=\bfseries
}

\title{Select-Then-Decompose: From Empirical Analysis to Adaptive Selection Strategy for Task Decomposition in Large Language Models}

\author{Shuodi Liu$^{*}$, Yingzhuo Liu$^{*}$, Zi Wang, Yusheng Wang,\\\textbf{Huijia Wu}, \textbf{Liuyu Xiang}$^{\dagger}$, \textbf{Zhaofeng He}\\  
  Beijing University of Posts and Telecommunications\\ \texttt{\{liushuodi, liuyingzhuo86, xiangly, zhaofenghe\} @bupt.edu.cn}
  }

\begin{document}
\maketitle
{
\let\thefootnote\relax\footnotetext{$^*$ Equal Contribution.}
\let\thefootnote\relax\footnotetext{
$^\dagger$ Corresponding authors. }
\let\thefootnote\relax\footnotetext{Accepted to the Main Conference of EMNLP 2025 (Oral).}
}

\input{latex/subfiles/0_abstract}
\input{latex/subfiles/1_intro}

\input{latex/subfiles/2_literature_review}
\input{latex/subfiles/3_insights}

\input{latex/subfiles/4_methodology}

\input{latex/subfiles/5_experiments}

\input{latex/subfiles/6_conclusion}
\input{latex/subfiles/7_limitation}

\bibliography{custom}

\newpage
\input{latex/subfiles/8_appendix}

\end{document}

%% file: latex/subfiles/0_abstract.tex
\begin{abstract}
Large language models (LLMs) have demonstrated remarkable reasoning and planning capabilities, driving extensive research into task decomposition. Existing task decomposition methods focus primarily on memory, tool usage, and feedback mechanisms, achieving notable success in specific domains, but they often overlook the trade-off between performance and cost. In this study, we first conduct a comprehensive investigation on task decomposition, identifying six categorization schemes. Then, we perform an empirical analysis of three factors that influence the performance and cost of task decomposition: categories of approaches, characteristics of tasks, and configuration of decomposition and execution models, uncovering three critical insights and summarizing a set of practical principles. Building on this analysis, we propose the \textit{Select-Then-Decompose} strategy, which establishes a closed-loop problem-solving process composed of three stages: selection, execution, and verification. This strategy dynamically selects the most suitable decomposition approach based on task characteristics and enhances the reliability of the results through a verification module. Comprehensive evaluations across multiple benchmarks show that the \textit{Select-Then-Decompose} consistently lies on the Pareto frontier, demonstrating an optimal balance between performance and cost. Our code is publicly available at \url{https://github.com/summervvind/Select-Then-Decompose}.
\end{abstract}

%% file: latex/subfiles/1_intro.tex
\section{Introduction}
Large Language Models (LLMs) have demonstrated excellent performance in the field of Natural Language Processing \cite{yang2024harnessing, chatgpt, gpt4, llama, llama2}. They have not only achieved remarkable success in basic tasks such as language understanding and text generation, but also exhibit strong reasoning and planning abilities \cite{qiao-etal-2023-reasoning, sun2023survey, chen2024llmarena, liu2025rainbowarena}. Motivated by these capabilities, researchers have increasingly focused on leveraging LLMs for task decomposition, allowing them to tackle complex problems in a step-by-step manner and improve overall accuracy and robustness \cite{kojima2022large, wang2023plan, sun2023pearl, yao2023react}.

Current research on task decomposition primarily enhances the performance of LLMs by integrating tool usage, feedback mechanisms, and memory modules \cite{chen2023autoagents, shen2023hugginggpt, zhang2025planning, qian-etal-2024-chatdev, li2024autokagglemultiagentframeworkautonomous}. However, there are still several questions that remain unanswered. For instance, what are the factors that influence performance and cost, and how can we balance the trade-off between them?

In this work, we investigate task decomposition in LLMs and introduce six categorization schemes: \ding{182} the interleaving sequence between decomposition and execution \cite{huang2024understandingplanningllmagents}, \ding{183} the number of LLM calls required to complete a task, \ding{184} the topological structure of decomposition, \ding{185} the format of decomposition, \ding{186} the range of subtask selection during decomposition, and \ding{187} whether tool usage is involved during execution. 

Based on these categorization schemes, we summarize five representative approaches for the following experiment and analysis. Subsequently, we conduct an in-depth empirical analysis to explore the main elements that influence the performance and cost of task decomposition, and identify three major contributing factors: categories of task decomposition approaches, characteristics of the tasks, configuration of the decomposition model and execution models. The experimental results reveal three important insights: \ding{182} The existing task decomposition approaches are confronted with a performance-cost dilemma; \ding{183} Task characteristics determine the sequence, calling form, and topology of task decomposition; \ding{184} Scaling the execution model yields greater performance gains than scaling the decomposition model, with the reasoning model further enhancing the execution stage. Based on these insights, we summarize a set of practical principles to guide task decomposition in LLMs, providing valuable insights for future research and practical deployment.

To further balance performance and cost, we propose the \textit{Select-Then-Decompose} strategy — a closed-loop framework composed of three collaborative modules: selection, execution, and validation. Instead of relying on a fixed decomposition paradigm, the selection module dynamically chooses the most suitable decomposition approach based on task complexity and characteristics. The execution module then applies the chosen approach to generate candidate solutions, while the validation module evaluates the confidence of the solutions and determines whether a fallback to a more sophisticated approach is necessary. Extensive experiments across diverse benchmarks demonstrate that \textit{Select-Then-Decompose} consistently lies on the Pareto frontier, striking an effective balance between performance and token cost.

Overall, our key contributions are: 
1) We provide a comprehensive investigation of task decomposition in LLMs, analyzing three key factors that impact its performance and cost, which lead to valuable insights and a set of practical principles; 2) We propose the \textit{Select-Then-Decompose} strategy, which dynamically selects appropriate decomposition approaches to mitigate the performance–cost dilemma; 3) Experimental results validate the superiority of the \textit{Select-Then-Decompose} strategy in multiple tasks, achieving an effective balance between task performance and token cost.

%% file: latex/subfiles/2_literature_review.tex

\section{Categorization of Task Decomposition Approaches}
\label{sec:2}
\begin{figure*}[t]  
    \centering
    \includegraphics[width=\textwidth]{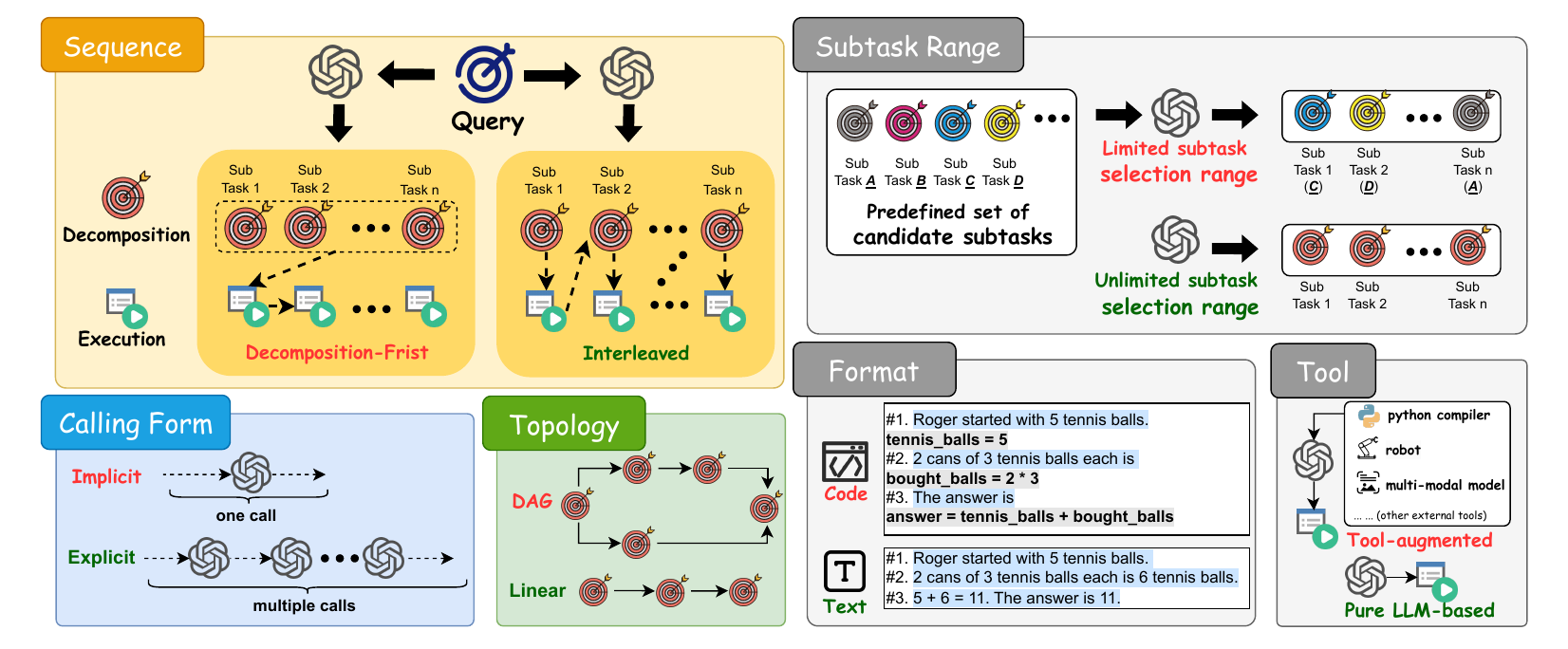}
    \caption{Task decomposition approaches are categorized from six perspectives, with those highlighted in color representing the primary focus of this study, while the categories in gray are not within the scope of our investigation.}
    \label{fig:2.1}
\end{figure*}

We systematically categorize existing task decomposition approaches and introduce six categorization schemes, with each scheme illustrated by a corresponding diagram in Figure~\ref{fig:2.1}.

\paragraph{Decomposition-First vs. Interleaved}

Task decomposition approaches generally consist of two main stages: decomposition and execution. Based on the interplay and sequence between these two stages, existing approaches can be classified into \textit{decomposition-first} approach \cite{shen2023hugginggpt, singh2023progprompt, sun2023pearl} and \textit{interleaved} approach \cite{yao2023react, wu2023visual, khot2022decomposed}. The former first decomposes the original task into a set of subtasks, which are then executed sequentially. In contrast, the latter performs decomposition and execution in an interleaved manner—only one subtask is generated at a time, and the next subtask is determined based on the outcome of the current task’s execution.



\paragraph{Implicit vs. Explicit}

Since solving problems with LLMs often involves multi-step generation, the manner in which LLMs are invoked plays a crucial role. Based on the frequency of LLM invocations, existing task decomposition approaches can be broadly categorized into \textit{explicit} approach \cite{shen2023hugginggpt, singh2023progprompt, zhou2022least} and \textit{implicit} \cite{kojima2022large, wang2023plan, gao2023pal} approach. The \textit{explicit} approach entails multiple LLM calls to separately carry out task decomposition and execution. In contrast, the \textit{implicit} approach seeks to integrate task understanding, decomposition, and execution within a single LLM invocation.



\paragraph{DAG vs. Linear}

In the decomposition stage, based on the dependency relationships between subtasks, the decomposition results can typically be classified into two common topological structures: \textit{linear} structure \cite{shen2023hugginggpt, singh2023progprompt, zhou2022least} and directed acyclic graph (\textit{DAG}) structure \cite{chen2023autoagents, wang2024dart, kannan2024smart}. In the \textit{linear} structure, a task is decomposed into a sequential chain of subtasks, where the output of each subtask directly serves as the input to the subsequent one. In contrast, the \textit{DAG} structure offers a more flexible and expressive decomposition paradigm, allowing for the parallel execution of independent subtasks and supporting complex dependency relationships, including both predecessors and successors.



\paragraph{Code vs. Text}

In the decomposition stage, the representation format of subtasks impacts the subsequent execution strategies and their effectiveness. Based on the design of current mainstream approaches, the decomposed subtasks are expressed in two common formats: \textit{code} format \cite{singh2023progprompt, kannan2024smart, gao2023pal} and \textit{text} format \cite{shen2023hugginggpt, zhou2022least, sun2023pearl}. In the \textit{code} format, subtask representations leverage structured languages (e.g., Python functions, JSON structures) to capture the execution logic and input-output dependencies between tasks. In contrast, the \textit{text format} describes each subtask’s content and objectives in natural language, offering greater flexibility and openness.



\paragraph{Limited Subtask Selection Range vs. Unlimited Subtask Selection Range}

In the decomposition stage, based on the size of the selectable subtask space, existing approaches can be classified into two categories: approaches with a \textit{limited subtask selection range} \cite{singh2023progprompt, sun2023pearl, wang2024dart} and approaches with an \textit{unrestricted subtask selection range} \cite{shen2023hugginggpt, zhou2022least, chen2023autoagents}. The former relies on a predefined set of candidate subtasks, from which the LLM must select when decomposing. These approaches allow for explicit control over the quality and relevance of the subtasks through the candidate set, providing stronger controllability and stability. In contrast, the latter do not depend on a predefined set of candidates. Instead, the LLM generates new subtasks freely based on the semantics and objectives of the task. 




\paragraph{Tool-Augmented vs. Pure LLM-based}

In the execution stage, existing approaches can be categorized according to whether external tools are involved in the execution of subtasks, resulting in two categories: \textit{tool-augmented} execution \cite{shen2023hugginggpt, singh2023progprompt,chen2023autoagents} and \textit{pure LLM-based} execution \cite{zhou2022least, sun2023pearl, khot2022decomposed}. \textit{Tool augmented} execution approaches leverage external tools-such as code interpreters, robots, or multi-modal models-to assist in completing specific subtasks. In contrast, \textit{pure LLM-based} execution approaches rely solely on the reasoning and generation capabilities of the LLM without invoking any external tools.





\paragraph{Summary}

Among the aforementioned six categorization schemes, the latter three schemes each contain a category that is typically tailored to specific tasks. For instance, the \textit{code} format is primarily applicable to domains such as robotic control and mathematical problem solving. Similarly, the \textit{limited subtask selection range} has limited applicability, as the set of candidate subtasks must be predefined in advance. In contrast, this study focuses on evaluating the effectiveness of various task decomposition approaches in general-purpose scenarios. Therefore, in the following sections, we concentrate on the categorization schemes introduced in the former three schemes, and select five representative approaches: CoT \cite{kojima2022large}, P\&S (Plan and solve) \cite{wang2023plan}, ReAct \cite{yao2023react}, P\&E (Plan and execute) \cite{sun2023pearl}, and P\&E with DAG structure \cite{sun2023pearl}, covering three categorization schemes. Table~\ref{tab:category} summarizes the specific categories of the five approaches. A comprehensive categorization taxonomy of all task decomposition approaches surveyed can be found in Appendix~\ref{appendix: taxonomy}. The details of each of the five representative approaches are described in Appendix~\ref{appendix: baselines}.


\begin{table*}[t]
\centering
\caption{Category assignments of the five representative methods}
\label{tab:category}
\resizebox{0.74\textwidth}{!}{%
\begin{tabular}{lccccc} 
\toprule
Categorization & COT & P\&S & ReAct & P\&E & P\&E (DAG) \\
\midrule
Decomposition-First(\ding{72}) vs. Interleaved(\ding{73}) & \ding{73} & \ding{72} & \ding{73} & \ding{72} & \ding{72} \\ 
Implicit(\ding{72}) vs. Explicit(\ding{73}) & \ding{72} & \ding{72} & \ding{73} & \ding{73} & \ding{73}\\ 
DAG(\ding{72}) vs. Linear(\ding{73}) & \ding{73} & \ding{73} & \ding{73} & \ding{73} & \ding{72}\\ 

\bottomrule
\end{tabular}
}
\end{table*}

%% file: latex/subfiles/3_insights.tex
\section{Empirical Analysis}
\label{sec:3}
While most existing research concentrates on applying decomposition approaches for designing task-specific workflows \cite{chen2023autoagents, zhang2025planning}, they often neglect to explore the underlying factors that fundamentally influence the performance and cost of task decomposition in LLMs. In this study, we conduct a systematic analysis of task decomposition from three perspectives: the performance–cost dilemma, the relationship between tasks and approaches, and the impact of model discrepancies. Based on extensive experiments and analysis, we present three key insights and summarize a set of practical principles.

\subsection{Performance-Cost Dilemma}
\label{sec:3.1}
To emphasize the performance and cost variations among different task decomposition approaches, we choose six approaches in total: IO (direct LLM invocation) and five representative approaches summarized in Section~\ref{sec:2}. Experiments are conducted on five widely used benchmarks: GSM8K \cite{cobbe2021gsm8k}, MATH \cite{hendrycks2measuring}, HumanEval \cite{chen2021codex}, Trivia Creative Writing \cite{wang-etal-2024-unleashing}, and HotpotQA \cite{yang-etal-2018-hotpotqa}. Detailed descriptions of these approaches and datasets are provided in the Appendix~\ref{appendix: Detailed Descriptions of Baselines and Datasets}. GPT-4o-mini \cite{openai_gpt4o_mini} is adopted as the base model, with the temperature parameter set to zero, and the beta parameter applied to fix the model seed at 42.\footnote{\url{https://platform.openai.com/docs/api-reference/chat/create}}

To compare the effectiveness of these approaches, we conduct five independent runs and report their mean performance and error bars across all five benchmarks in Table~\ref{tab:approaches}. IO shows limited effectiveness across tasks. In contrast, implicit approaches such as CoT and P\&S perform well on GSM8K, likely due to alignment with patterns encountered during post-training. However, their performance significantly degrades on more complex tasks like Trivia Creative Writing, which typically cannot be handled within a single LLM call. On the other hand, ReAct and P\&E achieve superior results on HumanEval and MATH. Notably, P\&E (DAG) delivers the best performance on Trivia Creative Writing and HotpotQA, and also performs competitively on the remaining benchmarks, resulting in the highest average score overall.

\begin{table*}[t]
\centering
\caption{Comparison of decomposition approaches across benchmarks.}
\label{tab:approaches}
\resizebox{\textwidth}{!}{%
\begin{tabular}{lccccccc}
\toprule
\multirow{2}{*}{Method} 
 & \multicolumn{2}{c}{Math} & \multicolumn{1}{c}{Code Generation} & \multicolumn{2}{c}{Creative Writing} & \multicolumn{1}{c}{Text Understanding} & \multirow{2}{*}{Avg.}\\ 
\cmidrule(lr){2-3} \cmidrule(lr){4-4} \cmidrule(lr){5-6} \cmidrule(lr){7-7}
 & GSM8K & MATH & HumanEval & Trivia Creative Writing(N=5) & Trivia Creative Writing(N=10) & HotpotQA \\
\midrule
IO          & 32.78 \textcolor{gray}{($\pm$0.15)} & 17.06 \textcolor{gray}{($\pm$0.21)} & 83.63 \textcolor{gray}{($\pm$0.27)} & 46.50 \textcolor{gray}{($\pm$0.36)} & 51.40 \textcolor{gray}{($\pm$0.41)} & 60.26 \textcolor{gray}{($\pm$0.24)} & 48.60 \\
CoT         & \textbf{93.45} \textcolor{gray}{($\pm$0.19)} & 50.11 \textcolor{gray}{($\pm$0.24)} & 86.20 \textcolor{gray}{($\pm$0.36)} & 49.58 \textcolor{gray}{($\pm$0.31)} & 51.34 \textcolor{gray}{($\pm$0.34)} & 63.21 \textcolor{gray}{($\pm$0.35)} & 65.65 \\
P\&S        & 92.12 \textcolor{gray}{($\pm$0.23)} & 49.40 \textcolor{gray}{($\pm$0.39)} & 84.54 \textcolor{gray}{($\pm$0.32)} & 48.82 \textcolor{gray}{($\pm$0.47)} & 51.24 \textcolor{gray}{($\pm$0.49)} & 62.06 \textcolor{gray}{($\pm$0.29)} & 64.70 \\
ReAct       & 91.56 \textcolor{gray}{($\pm$0.29)} & 44.68 \textcolor{gray}{($\pm$0.49)} & \textbf{89.85} \textcolor{gray}{($\pm$0.44)} & 61.48 \textcolor{gray}{($\pm$0.52)} & 62.72 \textcolor{gray}{($\pm$0.58)} & 53.28 \textcolor{gray}{($\pm$0.41)} & 67.26 \\
P\&E        & 92.47 \textcolor{gray}{($\pm$0.25)} & \textbf{52.13} \textcolor{gray}{($\pm$0.38)} & 82.46 \textcolor{gray}{($\pm$0.55)} & 62.50 \textcolor{gray}{($\pm$0.48)} & 54.26 \textcolor{gray}{($\pm$0.55)} & 63.04 \textcolor{gray}{($\pm$0.56)} & 67.81 \\
P\&E (DAG)  & 90.79 \textcolor{gray}{($\pm$0.39)} & 48.02 \textcolor{gray}{($\pm$0.37)} & 84.25 \textcolor{gray}{($\pm$0.30)} & \textbf{64.34} \textcolor{gray}{($\pm$0.49)} & \textbf{63.88} \textcolor{gray}{($\pm$0.57)} & \textbf{65.15} \textcolor{gray}{($\pm$0.48)} & \textbf{69.40} \\
\bottomrule
\end{tabular}
}
\end{table*}

\begin{figure*}[t]  
    \centering
    \includegraphics[width=\linewidth]{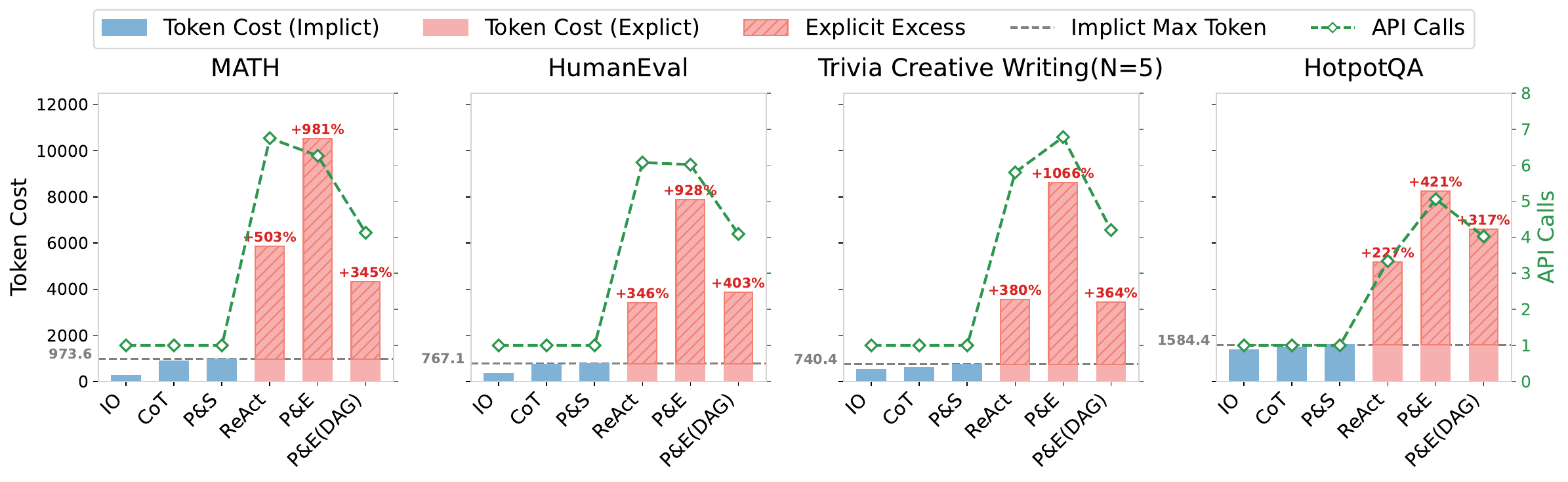}
    \caption{Token cost and API call analysis across benchmarks. The bar represents the token cost, and the line represents the API calls.}
    \label{fig:token_cost}
\end{figure*}

We also compare the average token consumption and API call frequency across the six approaches. Figure~\ref{fig:token_cost} presents the results for four representative benchmarks, with comprehensive results provided in Appendix~\ref{appendix: Original Data for Five Objective Benchmarks}. While explicit approaches achieve superior performance on certain benchmarks, they inevitably incur substantial costs in terms of token usage and API call frequency. For example, in the Trivia Creative Writing task (N=5), P\&E’s token consumption exceeds that of the implicit approach with the highest token usage by an astonishing $10 \times$, which is highly prohibitive. Although the P\&E (DAG) approach performs consistently well across all tasks, it still results in approximately $4 \times$ higher token consumption compared to implicit approaches.

\noindent\textbf{Takeaway I:} The existing task decomposition approaches are confronted with a performance-cost dilemma.

\subsection{The Relationship between Tasks and Approaches}
\label{sec:3.2}
Different studies have adopted diverse decomposition strategies to cope with domain-specific challenges. This has led us to a deeper exploration of the relationship between task characteristics and decomposition approaches.

Based on the results in Table~\ref{tab:approaches}, we observe that for mathematical tasks, CoT and P\&E exhibit the best performance. For code generation tasks, ReAct is the only one that achieves a score closest to 90. In writing and text comprehension tasks, P\&E (DAG) clearly outperforms other approaches. 
We hypothesize that these differences mainly stem from the distinct characteristics of each task type. To validate this hypothesis, we further conduct experiments on the MT-bench benchmark \cite{zheng2023judging}, with a detailed dataset setup described in Appendix~\ref{appendix: datasets}. The model and parameter settings are consistent with Section~\ref{sec:3.1}.

The experimental results, as shown in Table~\ref{tab:mt_bench}, indicate that for mathematical and reasoning tasks, CoT achieved the highest subjective scores. For writing and role-playing text tasks, P\&E (DAG) continued to excel, maintaining a leading position. However, in code generation tasks, ReAct, while ranking second, scored lower than CoT. We attribute this to the absence of a code execution component, which likely impacted ReAct's performance. Overall, this experiment not only mitigates metric inconsistencies across different benchmarks, but—through a unified subjective evaluation—also reinforces our hypothesis concerning the relationship between tasks and approaches. See the Appendix~\ref{appendix: Original Data for MT-Bench} for complete data on MT-bench.

\begin{table*}[ht]
\centering
\caption{MT-bench evaluation results: turn 1, turn 2, and average scores across approaches and five task categories, evaluated by Claude-3.5-Sonnet \cite{claude} as the judge and using GPT-4o-mini as the base model. \underline{Underlined} values indicate the highest score in each turn, while \textbf{bold} values indicate the highest average score.}

\label{tab:mt_bench}
\resizebox{\textwidth}{!}{%
\begin{tabular}{lccccccccccccccc}
\toprule
\multirow{2}{*}{Method} & \multicolumn{15}{c}{MT-bench} \\
\cmidrule{2-16}
 & \multicolumn{3}{c}{Writing} & \multicolumn{3}{c}{Roleplay} & \multicolumn{3}{c}{Reasoning} & \multicolumn{3}{c}{Math} & \multicolumn{3}{c}{Coding} \\
\cmidrule(lr){2-4} \cmidrule(lr){5-7} \cmidrule(lr){8-10} \cmidrule(lr){11-13} \cmidrule(lr){14-16}
 & Turn 1 & Turn 2 & Avg & Turn 1 & Turn 2 & Avg & Turn 1 & Turn 2 & Avg & Turn 1 & Turn 2 & Avg & Turn 1 & Turn 2 & Avg \\
\midrule
IO & 8.92 & 7.63 & 8.28 & 8.28 & 6.96 & 7.62 & 5.85 & 6.05 & 5.95 & 4.75 & 3.87 & 4.31 & 5.15 & 4.50 & 4.83 \\
CoT & 9.06 & 7.38 & 8.22 & 8.50 & 6.98 & 7.74 & \underline{8.55} & \underline{7.66} & \textbf{8.11} & \underline{9.73} & \underline{8.55} & \textbf{9.14} & \underline{7.82} & 5.69 & \textbf{6.76} \\
P\&S & 8.91 & \underline{8.03} & 8.47 & \underline{8.90} & 7.06 & 7.98 & 7.61 & 7.00 & 7.31 & 9.63 & 7.93 & 8.78 & 6.25 & 5.90 & 6.08 \\
ReAct & 8.10 & 7.98 & 8.04 & 6.77 & 6.23 & 6.50 & 8.08 & 7.12 & 7.60 & 9.58 & 7.32 & 8.45 & 6.00 & \underline{6.70} & 6.35 \\
P\&E & 9.05 & 7.04 & 8.05 & 8.53 & 6.39 & 7.46 & 8.13 & 6.02 & 7.08 & 9.33 & 7.70 & 8.52 & 4.95 & 5.27 & 5.11 \\
P\&E (DAG) & \underline{9.27} & 7.78 & \textbf{8.53} & 8.80 & \underline{7.77} & \textbf{8.29} & 8.08 & 6.83 & 7.46 & 9.47 & 7.50 & 8.49 & 4.86 & 4.87 & 4.87 \\

\bottomrule
\end{tabular}
}
\end{table*}

We then conduct a qualitative analysis to examine the relationship between task characteristics and approach categories, such as sequence, calling form, and topology. The logical rigor of math and reasoning tasks makes them well-suited to CoT's \textit{\textless implicit \textgreater} and P\&E's \textit{\textless linear \textgreater} strategies, both of which emphasize coherent, stepwise reasoning.
In contrast, the divergent thinking required by writing and comprehension tasks aligns with P\&E (DAG)'s parallel decomposition. Code generation, characterized by iterative refinement, benefits more from ReAct's \textit{\textless explicit, interleaved \textgreater} strategy. These findings suggest that task-specific cognitive demands fundamentally influence the suitability of decomposition approaches.

\noindent\textbf{Takeaway II:} Task characteristics determine the sequence, calling form, and topology of task decomposition.

\subsection{Impact of Model Discrepancies}
The parameter scale and reasoning capability of a model are key factors influencing its performance \cite{kaplan2020scaling, shao2024deepseekmath}. We similarly focus on the specific roles these factors play in the stages of decomposition and execution. Based on the discussion in Section~\ref{sec:3.1}, we select the P\&E (DAG) approach as the research carrier, primarily due to its explicit decomposition pattern and the requirement of a structured plan. Systematic experiments are conducted on the MATH dataset.

\begin{figure}[t]  
    \centering
    \includegraphics[width=\columnwidth]{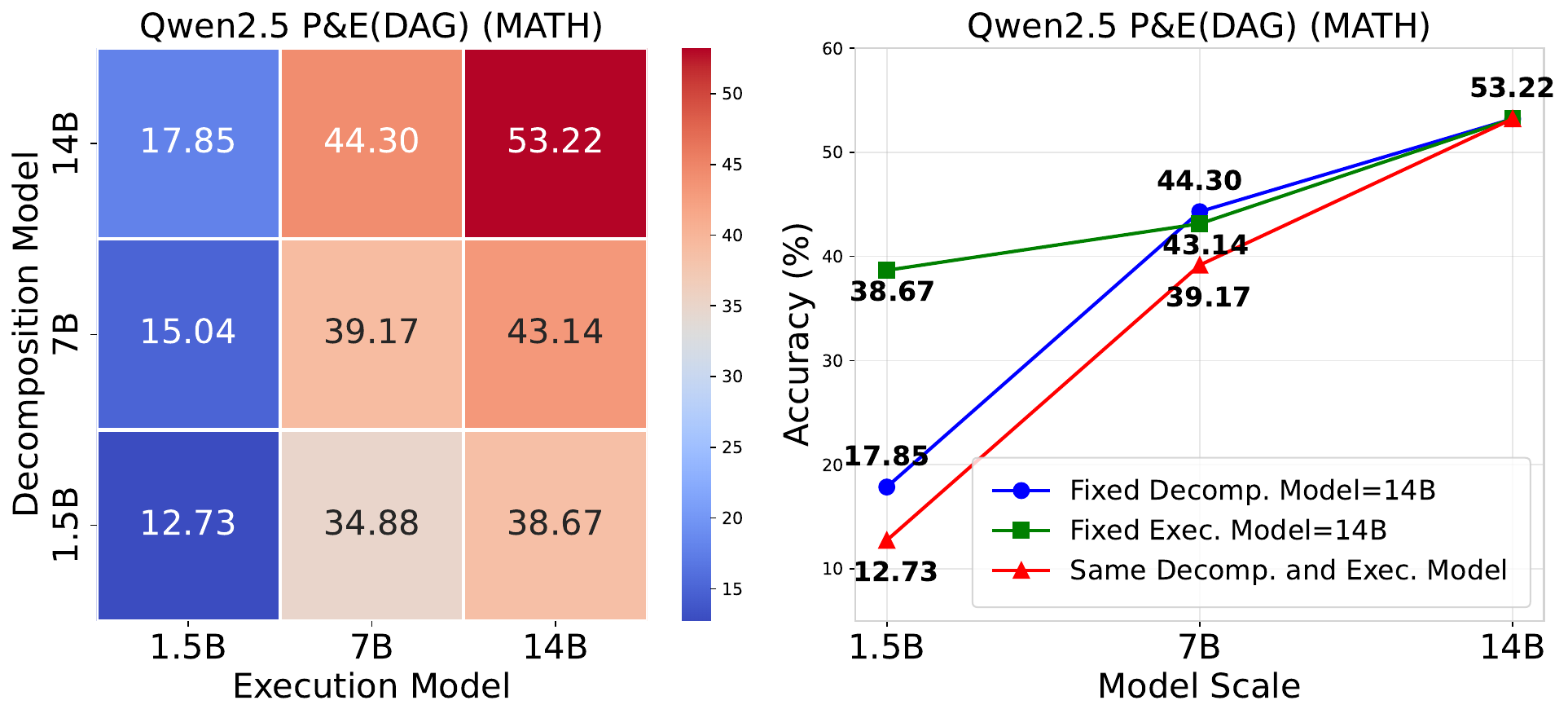}
    \caption{The left panel presents nine cross-model experiments, while the right panel shows three sets of controlled experiments.}
    \label{fig:cross_models}
\end{figure}

First, we explore the impact of model scale. We use three language models from the Qwen2.5 series \cite{yang2024qwen2} with different parameter sizes (Qwen2.5-1.5B/7B/14B-instruct) as the decomposition and execution models, forming nine cross-model experiment combinations. As shown in the left panel of Figure~\ref{fig:cross_models}, from both the decomposition and execution model perspectives, overall accuracy improves with larger model parameter sizes, indicating that model scale has a positive impact on performance.

To further analyze the individual and comparative impacts of the decomposition model and the execution model, we design three sets of controlled experiments, as shown in the right panel of Figure~\ref{fig:cross_models}. 
We observe that the performance of the execution model has a more significant impact on overall task performance, as evidenced by two key findings: First, the slope of the blue line is noticeably steeper than that of the green line, indicating that performance improvements from scaling the execution model size far exceed those from scaling the decomposition model. Second, the blue line closely aligns with the red line, confirming that the adjustment in the execution model is the primary driver behind the scaling law phenomenon. 

\begin{figure}[t]  
    \centering
    \includegraphics[width=\columnwidth]{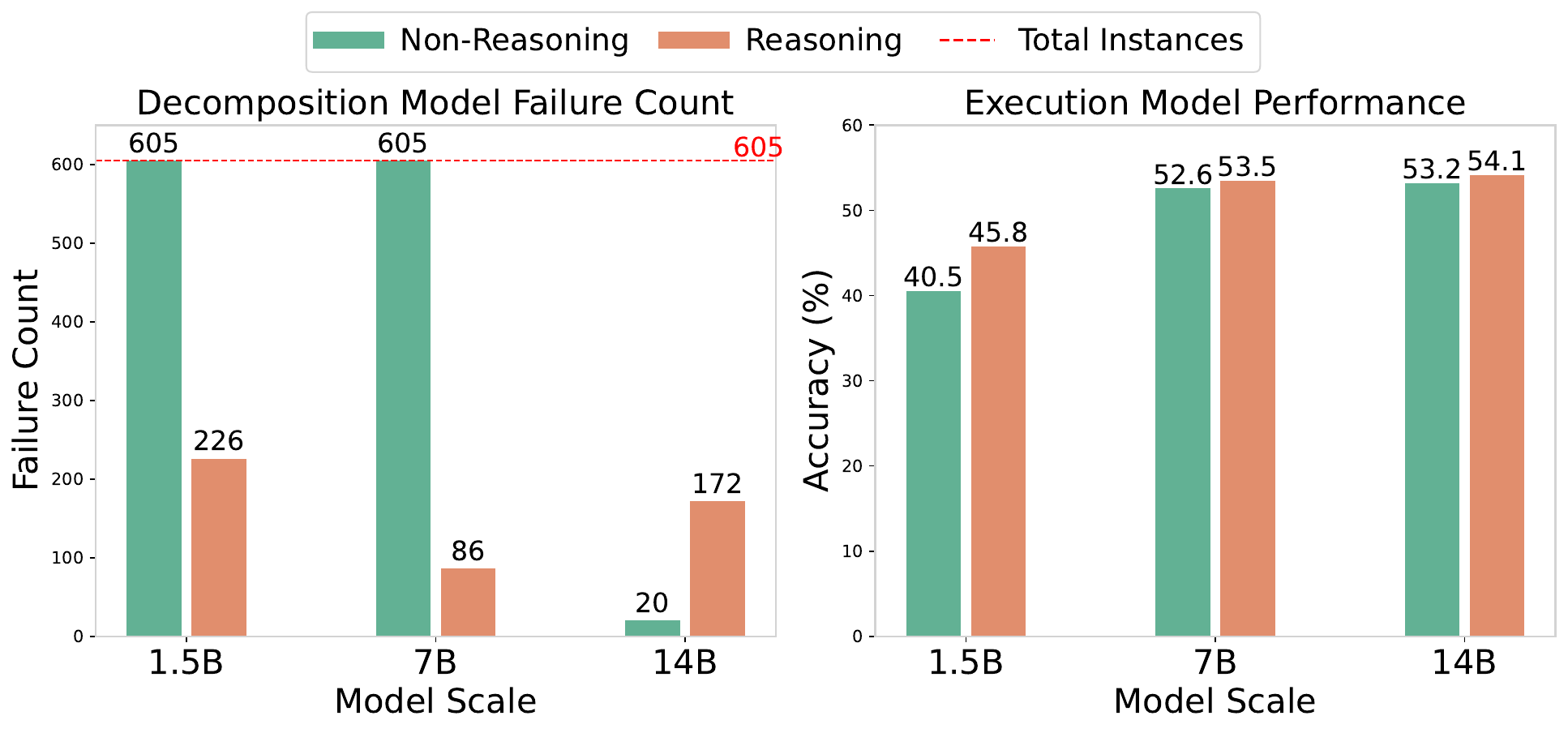}
    \caption{Performance comparison between reasoning and non-reasoning models in decomposition and execution stages.}
    \label{fig:reasoning_model}
\end{figure}

In addition, we examine the impact of reasoning capability by comparing the performance of models (Qwen2.5-math-1.5B/7B-instruct, Qwen2.5-14B-instruct) with their corresponding Deepseek-R1 distilled versions \cite{deepseekai2025deepseekr1incentivizingreasoningcapability} as decomposition models and execution models. Firstly, we report the number of invalid plans generated during the decomposition stage (left panel of Figure~\ref{fig:reasoning_model}). The results show that non-reasoning models at both 1.5B and 7B scales struggle to generate plans that comply with the required format, while their distilled reasoning counterparts show marked improvements. However, at the 14B scale, reasoning models produce more invalid outputs, suggesting that increased reasoning ability may compromise format control, especially when abstract reasoning is prioritized over structural compliance. Secondly, we compare the execution-stage performance of reasoning and non-reasoning models, using Qwen2.5-14B-instruct as a fixed decomposition model (right panel of Figure~\ref{fig:reasoning_model}). The results show that reasoning models consistently outperform their non-reasoning counterparts across all parameter scales, highlighting their advantages in the execution stage. 


\noindent\textbf{Takeaway III:} Scaling the execution model yields greater performance gains than scaling the decomposition model, with the reasoning model further enhancing the execution stage.

\subsection{Practical Principles}

Based on the above experiments and analysis, we summarize a set of practical principles to guide the selection of appropriate task decomposition approaches and models, providing actionable guidance for real-world applications.

The practical principles can be formulated as follows: Firstly, when approaching a question, one should choose a decomposition approach that matches the task characteristics, such as logical rigor, divergence, or iterativity. For example, CoT is suitable for mathematical problems, P\&E (DAG) for writing tasks, and ReAct for coding tasks. Secondly, the choice of models should depend on resource constraints and performance requirements. If an \textit{implicit} decomposition method is adopted, selecting a single model with strong performance is sufficient. In contrast, for \textit{explicit} decomposition, it is recommended to use a model with strong performance for execution and a model with basic instruction-following ability for decomposition.

%% file: latex/subfiles/4_methodology.tex
\section{Methodology}

\subsection{Select-Then-Decompose Strategy}
In addition to a set of practical principles, we also endeavor to optimize the balance between task performance and cost. 
Based on the insights in Section~\ref{sec:3.1} and Section~\ref{sec:3.2}, we propose a novel and efficient strategy called \textit{Select-Then-Decompose} (S\&D), which dynamically selects an appropriate decomposition approach according to the task's complexity and characteristics. This allows for achieving an optimal balance between performance and cost.

The \textit{Select-Then-Decompose} strategy mainly consists of three functional modules: the \textbf{Selection Module}, the \textbf{Execution Module}, and the \textbf{Validation Module}. These three modules work collaboratively to form a closed-loop task-solving process. The detailed algorithmic procedure is described in Appendix~\ref{appendix: algorithm}.

The \textbf{Selection Module} is the core component of the \textit{Select-Then-Decompose} strategy. Powered by an LLM, this module employs a carefully designed set of prompts \( P \) to guide the LLM in analyzing and understanding the input question \( Q \), and returns the most suitable decomposition approach \( A \) along with the reasoning \( R \).
The complete prompt for the Selection Module is provided in the Appendix~\ref{appendix: Prompt for Select-Then-Decompose Strategy}.

The \textbf{Execution Module} follows the approach $A$ selected by the Selection Module and applies the corresponding decomposition algorithm to the input task $Q$, generating a candidate solution $S$. The Execution Module is designed with high modularity and extensibility.

The \textbf{Validation Module} leverages an LLM to assess the confidence score \( C \in [0, 1] \) of the candidate solution \( S \), based on the original question \( Q \). If \( C \geq T \), where \( T \) is a predefined threshold, the solution is accepted; otherwise, the system initiates a staged switching mechanism that sequentially explores \{IO\}, implicit approaches \{\text{CoT}, \text{P\&S}\}, and explicit approaches \{\text{ReAct}, \text{P\&E}, \text{P\&E (DAG)}\}. Within each category, the method is selected via uniform random sampling \( (M \sim \mathcal{U}(G)) \). Detailed prompting instructions are provided in the Appendix~\ref{appendix: Prompt for Select-Then-Decompose Strategy}.

%% file: latex/subfiles/5_experiments.tex
\section{Experiments}
\subsection{Setup}
The baselines and benchmarks in our experiments follow the same settings as those described in Section~\ref{sec:3.1}. Additionally, S\&D strategy employs a validation threshold of 0.7 and allows up to 3 switching iterations. 

\subsection{Main Results}

\begin{figure*}[t]  
    \centering
    \includegraphics[width=\linewidth]{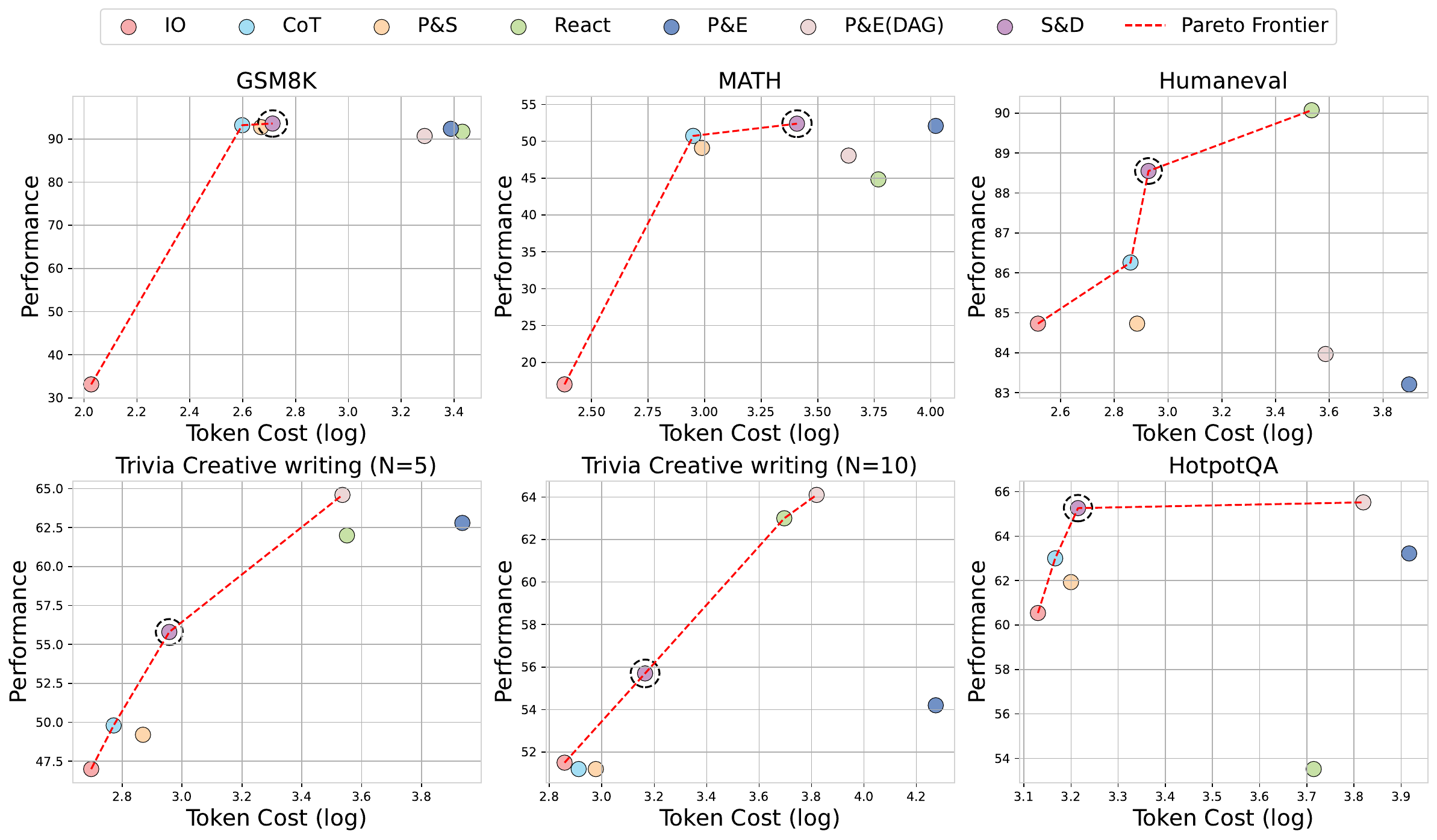}
   \caption{Performance vs. cost trade-offs across benchmarks}
    \label{fig:pareto_scatter}
\end{figure*}

The main experimental results as shown in Figure~\ref{fig:pareto_scatter}. S\&D consistently lies on the Pareto frontier across five benchmark tasks, demonstrating a favorable balance between performance and cost. The complete raw data is in the Appendix~\ref{appendix: Original Data for Five Objective Benchmarks}. 
Notably, on tasks where candidate approaches show small performance gaps, such as GSM8K and MATH, S\&D achieves higher accuracy with minimal additional cost. This advantage stems from its LLM-based selection mechanism, which can identify the appropriate approach according to the question, thus outperforming any individual method. On the Humaneval and HotpotQA datasets, S\&D attains near-optimal performance while using only 24.77\% of the average token cost, achieving Pareto optimality. In the Trivia Creative Writing task, where implicit and explicit approaches exhibit significant performance differences, S\&D demonstrates an approximately linear trade-off between performance and cost along the Pareto frontier.

\begin{figure}[t]  
    \centering
    \includegraphics[width=\linewidth]{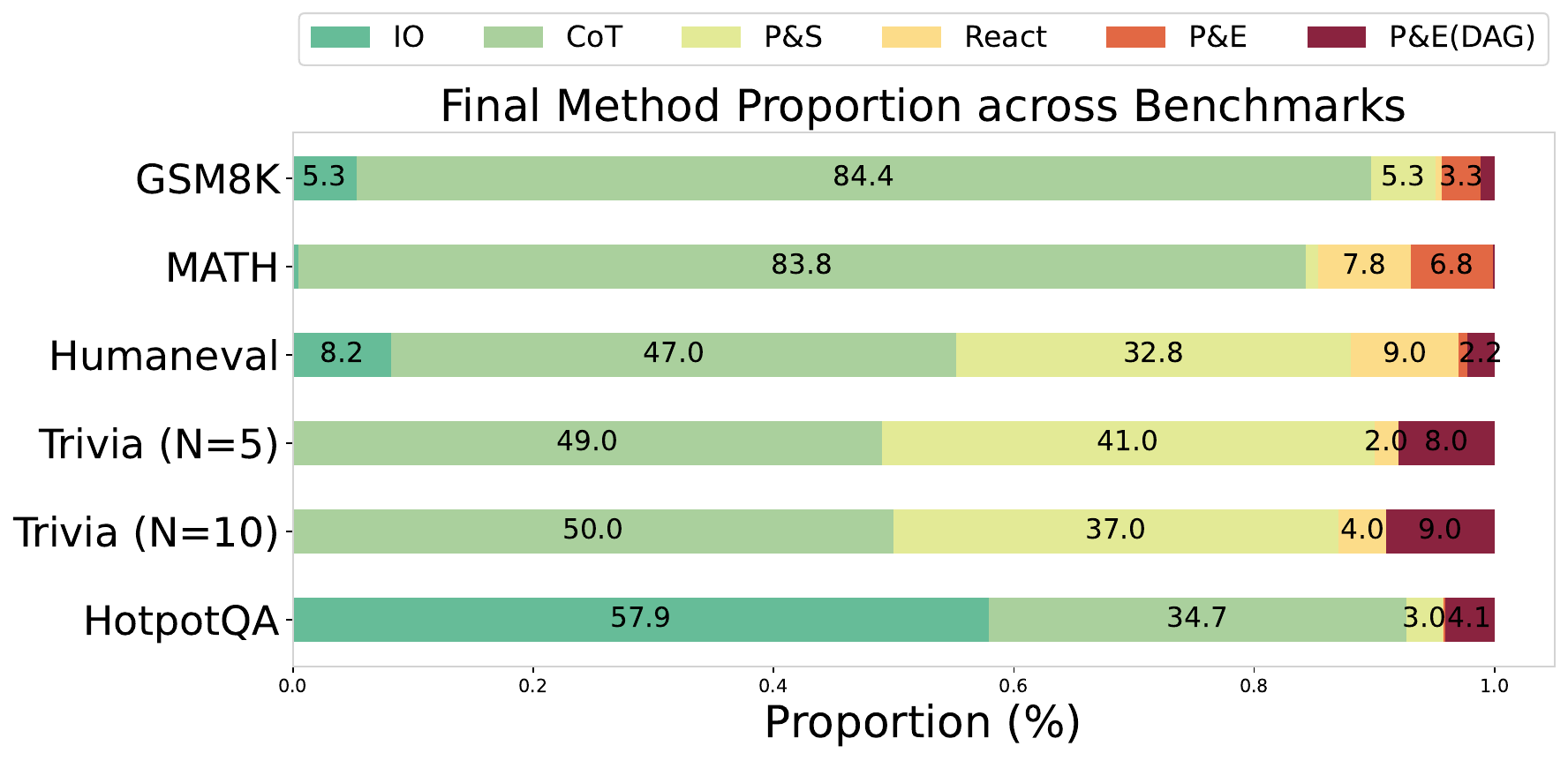}
    \caption{Proportional distribution of final approaches selected by the \textit{Select-Then-Decompose} strategy across different benchmarks}
    \label{fig:Final_Method_Proportion}
\end{figure}

We further analyze the proportions of final approaches selected by the S\&D strategy across different benchmarks after the “Select-Execute-Verify” process, as shown in Figure~\ref{fig:Final_Method_Proportion}.  Overall, Implicit decomposition approaches dominate, comprising approximately 85\%, while explicit approaches account for only about 15\%. This suggests that S\&D favors low-cost implicit strategies and only switches to explicit methods mainly for complex problems or failed verifications. Task-specific patterns reveal that, relative to other tasks, CoT and P\&E are more common in mathematical tasks, ReAct sees greater usage in code generation, and P\&E (DAG) is more prevalent in writing and text understanding. These findings align with the insights in Section~\ref{sec:3.2}, highlighting the relationship between tasks and decomposition approaches. See the Appendix~\ref{appendix: examples} for more examples.


\subsection{Analysis}
\paragraph{Ablation Study}

To assess the contributions of the selection and verification modules in the S\&D strategy, we perform an ablation study on the key components of the S\&D strategy using the GPT-4o-mini model on HumanEval: (1) \textbf{w/o Select (IO)}, removing the selection module and using IO as the initial approach; (2) \textbf{w/o Select (Random)}, removing the selection module and using a random initial approach as the initial approach; and (3) \textbf{w/o Val.}, removing the verification module. As shown in Table~\ref{tab:ablation_experiment}, removing the selection module and using IO as the initial approach reduces token usage but degrades performance, while using a random initial approach increases token consumption and also lowers accuracy. This suggests that the selection module enhances efficiency and effectiveness by guiding the choice of decomposition. Additionally, omitting the verification module leads to a significant performance drop, underscoring its role in mitigating hallucinations and improving solution reliability.

\begin{table}[htbp]
\centering
\resizebox{\columnwidth}{!}{
\begin{tabular}{lccc}
\toprule
\textbf{Config.} & \textbf{Performance} & \textbf{Avg. Token Cost} \\
\midrule
w/ Select \& Val. & 88.55 & 845.82 \\
\midrule
w/o Select (IO) & 86.59 (\textcolor{red}{↓2.21\%}) & 542.35 (\textcolor{green}{↓35.89\%})\\
w/o Select (Random) & 87.19 (\textcolor{red}{↓1.53\%}) & 2782.34 (\textcolor{red}{↑228.99\%})  \\
w/o Val. & 85.98 (\textcolor{red}{↓2.90\%}) & 753.29 (\textcolor{green}{↓10.94\%}) \\
\bottomrule
\end{tabular}
}
\caption{Ablation study for \textit{Select-Then-Decompose} strategy on HumanEval. Percentage changes are relative to the full configuration.}
\label{tab:ablation_experiment}
\end{table}



\paragraph{Sensitivity Analysis}

\begin{figure}[htbp]
    \centering
    \includegraphics[width=\linewidth]{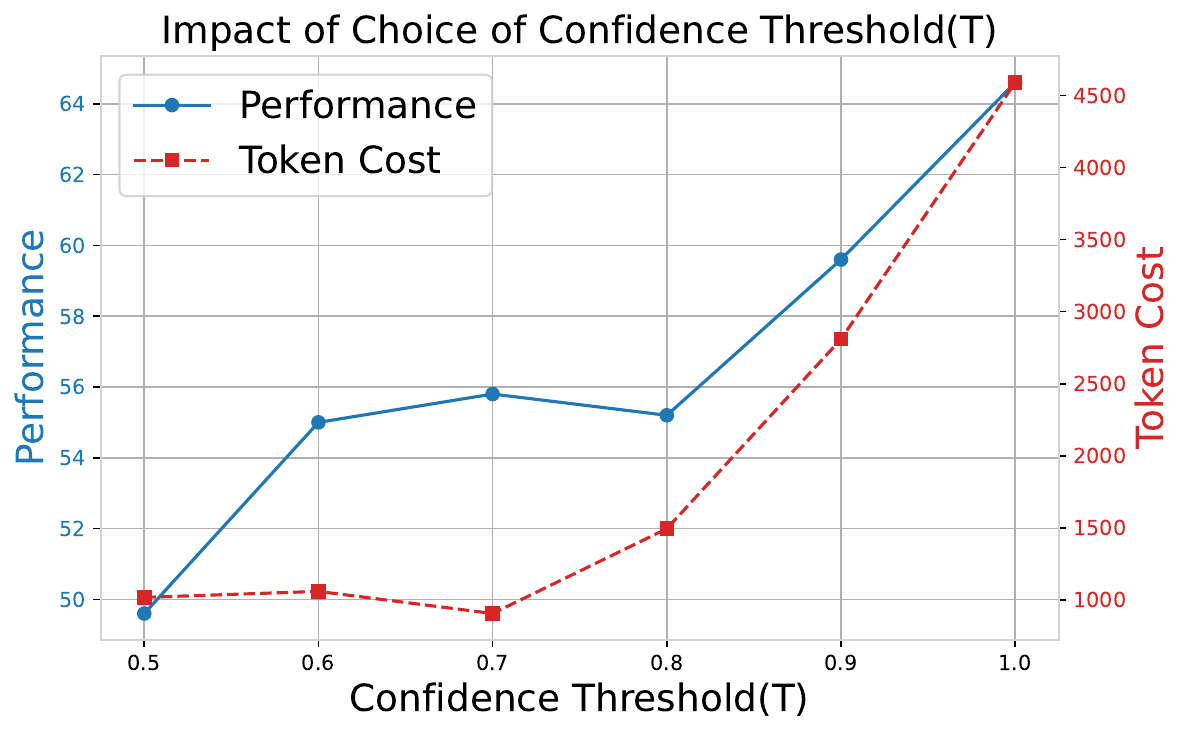}
    \caption{Impact of the choice of confidence threshold $T$ on the Trivia Creative Writing dataset.}
    \label{fig:threthold}
\end{figure}

We further investigate the effect of the confidence threshold $T$ through experiments on the Trivia Creative Writing dataset. As shown in Figure~\ref{fig:threthold}, raising the threshold notably enhances system performance by filtering out unreliable candidate solutions. However, this improvement comes at the cost of increased token consumption, particularly beyond the 0.9 threshold. From Figure~\ref{fig:threthold}, it can be observed that when the threshold is set to 0.7, the model achieves a good balance between performance and cost: the token cost is minimized within the range of 0.5 to 1.0, and the resulting performance even exceeds that of a threshold of 0.8, which motivates our choice of $T=0.7$ as the default threshold. Of course, one can choose a larger threshold to achieve slightly better performance, but this inevitably leads to higher token costs.

\paragraph{Generalization Analysis}

\begin{figure}[htbp]
    \centering
    \includegraphics[width=\linewidth]{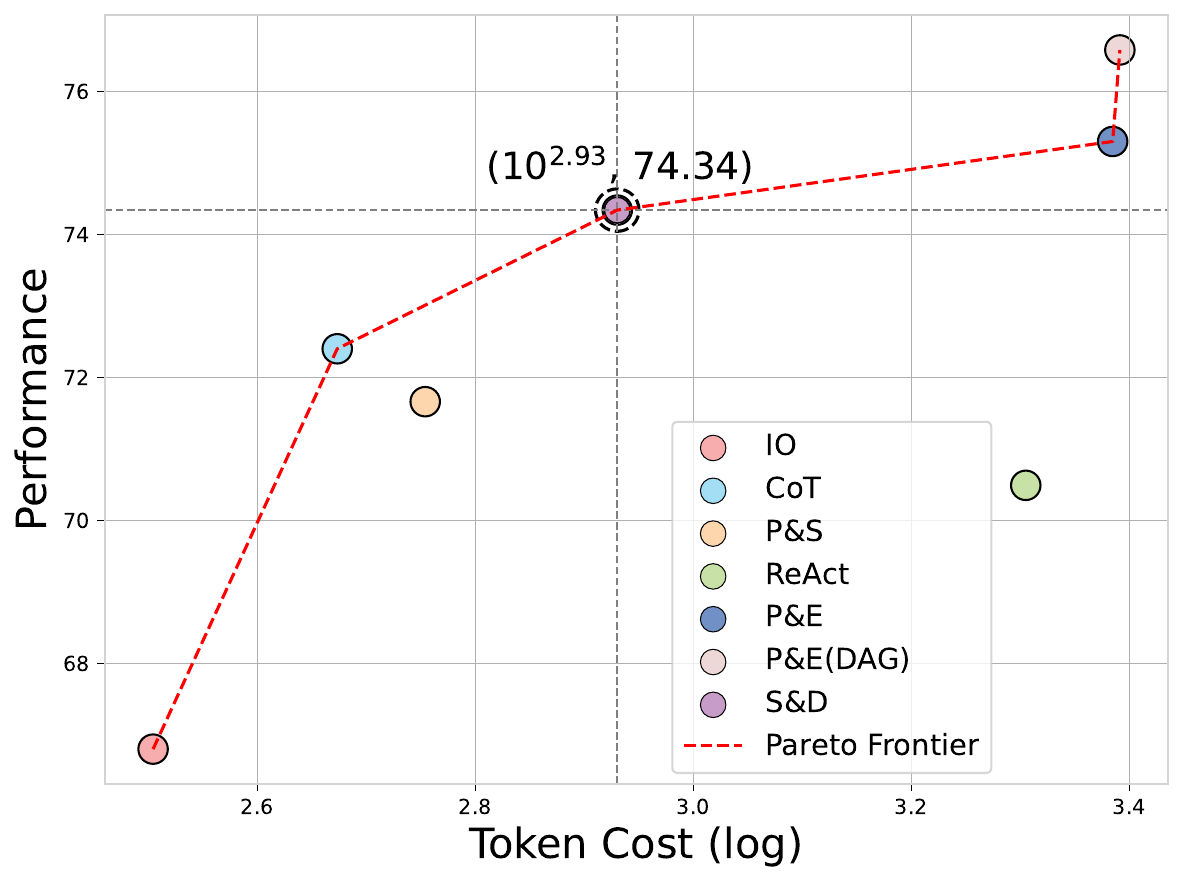}
    \caption{Generalization experiment on DROP.}
    \label{fig:DROP}
\end{figure}

The prompt for the selection module (see Appendix~\ref{appendix: Prompt for Select-Then-Decompose Strategy}) is designed based on the observations in Section~\ref{sec:3}, excluding any task- or dataset-specific information. To further demonstrate the generalizability of our method, we conduct additional experiments on the DROP benchmark \cite{dua2019drop}. As shown in Table~\ref{fig:DROP}, the results indicate that the \textit{Select-Then-Decompose} strategy still maintains a Pareto advantage. We will discuss scenarios when S\&D fails or performs poorly in the limitations section.

%% file: latex/subfiles/6_conclusion.tex
\section{Conclusion}
In conclusion, our research explores the task decomposition in LLMs. We first investigate existing task decomposition research into six categorization schemes and identify five representative approaches. Through experiments and analysis across approaches, tasks, and models, we present three insights into task decomposition and propose a set of practical principles to guide real-world applications. In addition, the issue of the high cost of existing methods leads us to propose the \textit{Select-Then-Decompose} strategy, which dynamically chooses suitable decomposition approaches based on the task. Extensive experiments show that our strategy consistently lies on the Pareto frontier, achieving a strong balance between performance and cost. Our contributions not only enhance the understanding of LLM task decomposition and offer a practical framework for balancing performance and cost.

%% file: latex/subfiles/7_limitation.tex
\section*{Limitations}

Although our research focuses on task decomposition in LLMs, we acknowledge two primary limitations. First, we only examined the decomposition mechanism, without exploring the representation formats (e.g., code or text) or the use of external tools, both of which have been shown in prior work to improve performance.  Thus, for tasks that require the use of specialized tools,  the S\&D strategy might need to be adapted accordingly in order to function properly. Second, our S\&D strategy relies solely on prompting the model to choose a suitable decomposition approach based on the task, without any additional training to enhance this capability. So if the model used in the selection module is relatively weak, its quality may degrade, leading to poor overall performance. We encourage future research in these two promising directions to further advance our understanding of autonomous task decomposition in LLMs.

\section*{Ethical Considerations}
While task decomposition in LLMs offers significant advancements in tackling complex problems efficiently, it also raises important ethical concerns. The increased use of LLMs with dynamic decomposition strategies can lead to unintended consequences such as over-reliance on automated decision-making, potential biases inherited from training data, and privacy risks when handling sensitive information. Additionally, the token cost and computational resources required may contribute to environmental impacts and raise accessibility issues for smaller organizations or communities. Therefore, it is crucial to design task decomposition approaches with transparency, fairness, and sustainability in mind, ensuring that these technologies are deployed responsibly and inclusively.

\section*{Acknowledgements}
This work was supported in part by the National Natural Science Foundation of China (Nos. 62176025, 62576046, 62301066, U21B2045, 62206012, and 62406028), the Fundamental Research Funds for the Central Universities (No. 2023RC72), the Key Project of Philosophy and Social Sciences Research, Ministry of Education, China (No. 24JZD040), and the Chinese Nutrition Society (No. CNS-YUM2024-120).

%% file: latex/subfiles/8_appendix.tex
\appendix
\definecolor{darkorange}{RGB}{255, 140, 0}
\definecolor{darkblue}{RGB}{84, 112, 198}
\definecolor{lightgreen}{RGB}{145, 204, 117}
\definecolor{lightyellow}{RGB}{250, 200, 88}
\definecolor{lightred}{RGB}{238, 102, 102}
\definecolor{lightblue}{RGB}{115, 192, 222}

\newtcolorbox{promptbox}[2][Prompt]{
colback=black!5!white,
arc=5pt, 
boxrule=0.5pt,
fonttitle=\bfseries,
title=#1, 
before upper={\small}, fontupper=\fontfamily{ptm}\selectfont,
colframe=#2, 
}

\section{Taxonomy of Task Decomposition Methods}

We categorize task decomposition methods from six distinct perspectives, summarizing the categorization of existing approaches accordingly. Specifically, "Perspective 1" corresponds to "Decomposition-First vs. Interleaved", "Perspective 2" to "Implicit vs. Explicit", "Perspective 3" to "DAG vs. Linear", "Perspective 4" to "Code vs. Text", "Perspective 5" to "Limited Subtask Selection Range vs. Unlimited Subtask Selection Range", and "Perspective 6" to "Tool-Augmented vs. Pure LLM-based". The symbol "\ding{72}" denotes the former category in each perspective, while "\ding{73}" denotes the latter.

\label{appendix: taxonomy}

\begin{table*}[htbp]
\centering
\caption{Category assignments of the five representative methods}
\label{tab:taxonomy category}
\resizebox{\textwidth}{!}{%
\begin{tabular}{l|cccccc} 
\toprule
Approach & Perspective 1 & Perspective 2 & Perspective 3 & Perspective 4 & Perspective 5 & Perspective 6\\
\midrule
HuggingGPT \cite{shen2023hugginggpt} & \ding{72} & \ding{73} & \ding{73} & \ding{73} & \ding{73} & \ding{72}\\ 

ProgPrompt \cite{singh2023progprompt} & \ding{72} & \ding{73} & \ding{73} & \ding{72} & \ding{72} & \ding{72}\\ 

Least-to-most \cite{zhou2022least} & \ding{72} & \ding{73} & \ding{73} & \ding{73} & \ding{73} & \ding{73}\\ 

PEARL \cite{sun2023pearl} & \ding{72} & \ding{73} & \ding{73} & \ding{73} & \ding{72} & \ding{73}\\ 

AutoAgents \cite{chen2023autoagents} & \ding{72} & \ding{73} & \ding{72} & \ding{73} & \ding{73} & \ding{72}\\ 

DART-LLM \cite{wang2024dart} & \ding{72} & \ding{73} & \ding{72} & \ding{73} & \ding{72} & \ding{72}\\ 

SMART-LLM \cite{kannan2024smart} & \ding{72} & \ding{73} & \ding{72} & \ding{72} & \ding{72} & \ding{72}\\ 

ReAct \cite{yao2023react} & \ding{73} & \ding{73} & \ding{73} & \ding{73} & \ding{73} & \ding{72}\\ 

Visual ChatGPT \cite{wu2023visual} & \ding{73} & \ding{73} & \ding{73} & \ding{73} & \ding{73} & \ding{72}\\ 

Decomposed Prompting \cite{khot2022decomposed} & \ding{73} & \ding{73} & \ding{73} & \ding{73} & \ding{73} & \ding{73}\\ 

Task Navigator \cite{ma2024task} & \ding{73} & \ding{73} & \ding{73} & \ding{73} & \ding{73} & \ding{73}\\ 

Plan-and-Solve \cite{wang2023plan} & \ding{72} & \ding{72} & \ding{73} & \ding{73} & \ding{73} & \ding{73}\\ 

CoT \cite{kojima2022large} & \ding{73} & \ding{72} & \ding{73} & \ding{73} & \ding{73} & \ding{73}\\ 

PAL \cite{gao2023pal} & \ding{73} & \ding{72} & \ding{73} & \ding{72} & \ding{73} & \ding{72}\\ 

PoT \cite{chen2022program} & \ding{73} & \ding{72} & \ding{73} & \ding{72} & \ding{73} & \ding{72}\\ 

TDAG \cite{wang2025tdag} & \ding{73} & \ding{72} & \ding{73} & \ding{73} & \ding{73} & \ding{73}\\ 

ADaPT \cite{prasad2023adapt} & \ding{73} & \ding{72} & \ding{73} & \ding{73} & \ding{73} & \ding{73}\\ 

Re-TASK \cite{wang2024re} & \ding{73} & \ding{72} & \ding{73} & \ding{73} & \ding{73} & \ding{73}\\ 

\bottomrule
\end{tabular}
}
\end{table*}

\section{Detailed Descriptions of Baselines and Datasets}
\label{appendix: Detailed Descriptions of Baselines and Datasets}
\subsection{Baselines}
\label{appendix: baselines}

We select five representative approaches: CoT, P\&S, ReAct, P\&E, and P\&E (DAG). The details of these approaches are as follows:

\begin{itemize}
    \item \textbf{CoT} \cite{kojima2022large}: CoT enables LLMs' zero-shot reasoning with the prompt "Let's think step-by-step."
    \item \textbf{P\&S} \cite{wang2023plan}: P\&S improves upon CoT by splitting it into two instructions: "Let's first devise a plan" and "Let's carry out the plan."
    \item \textbf{ReAct} \cite{wang2023plan}: Unlike CoT, which embeds reasoning within planning, ReAct alternates between reasoning (Thought) and acting (Action).
    \item \textbf{P\&E} \cite{sun2023pearl}: P\&E decomposes the task into a multi-step plan and executes each subtask sequentially.
    \item \textbf{P\&E (DAG)} \cite{sun2023pearl}: The key difference between P\&E (DAG) and P\&E is that the former produces a DAG-structured plan, while P\&E generates a linear-structured plan. 
\end{itemize}

\subsection{Benchmarks}
\label{appendix: datasets}
We select four representative task categories: reasoning, code generation, creative writing, and text comprehension. To ensure a comprehensive and multidimensional assessment, we adopt five objective benchmarks and one subjective benchmark. The details of tasks and benchmarks are as follows:
\paragraph{Objective Benchmarks} We employ five publicly available benchmarks with well-defined quantitative metrics covering four task categories.
\begin{itemize}[leftmargin=10pt]
\item \textbf{Reasoning Task.}
\textbf{GSM8K} \cite{cobbe2021gsm8k} provides a comprehensive set of elementary school-level word problems, designed to evaluate arithmetic reasoning capabilities. We assess the quality of generated solutions via accuracy (\%), with the full dataset for testing. 
\textbf{MATH} \cite{hendrycks2measuring} integrates high-difficulty mathematical competition problems covering seven mathematical fields, categorized into five difficulty levels. We similarly assess using the accuracy (\%) for measuring the quality of the generated solutions. Following \cite{hong2024datainterpreterllmagent}, we select 617 problems from four representative problem types (Combinatorics \& Probability, Number Theory, Pre-algebra, and Pre-calculus) at difficulty level 5.

\item \textbf{Code Generation Task.}
\textbf{HumanEval} \cite{chen2021codex} is a widely recognized function-level code generation benchmark,   tailored for assessing fundamental programming skills. We assess adopting the pass@k as a measure of function correctness across multiple standard test cases, with the full dataset for testing.

\item \textbf{Creative Writing Task.}
\textbf{Trivia Creative Writing} \cite{wang-etal-2024-unleashing} requires generating a coherent narrative based on a given topic while integrating answers from N(=5/10) trivia questions, designed to quantify the model’s information synthesis and writing abilities. We assess using an automatic metric score, calculated by the proportion of correct answer mentions. We use the full dataset consists of 100 instances each for N=5 and N=10, totaling 200 samples.

\item \textbf{Text Comprehension Task.}
\textbf{HotpotQA} \cite{yang-etal-2018-hotpotqa} integrates Wikipedia-based multi-hop question-answer pairs, designed to assess text comprehension abilities by requiring answers derived from multiple supporting documents. We assess via the F1 score, which quantifies the balance between precision and recall in identifying the correct answers. In line with prior works \cite{hu2025automateddesignagenticsystems, shinn2023reflexion}, we randomly select 1,000 samples for evaluation.
\textbf{DROP} \cite{dua2019drop} is an English reading comprehension benchmark. Unlike HotpotQA, DROP focuses more on discrete reasoning operations—such as addition, subtraction, and comparison—within a single paragraph.
\end{itemize}

\paragraph{Subjective Benchmark}
To mitigate the impact of evaluation differences caused by different benchmark indicators, we choose \textbf{MT-bench} \cite{zheng2023judging}, which provides 80 high-quality open-ended questions covering 8 task categories and is evaluated through models or human subjective scores. We asses using Claude-3.5-Sonnect as the evaluation model to rate the responses on a scale from 0 to 10. We selected 50 high-quality open-ended questions from MT-bench, corresponding to different task categories (mathematics, reasoning, coding, writing, and role-playing). Through this benchmark, we ensure consistency in evaluation metrics across tasks and effectively assess the performance of various decomposition methods in handling open-ended questions.


\section{Algorithm of Select-Then-Decompose Strategy}
\label{appendix: algorithm}

To balance task performance and computational cost, we propose the \textit{Select-Then-Decompose} (S\&D) strategy. 

The strategy comprises three key modules: \textbf{Selection}, \textbf{Execution}, and \textbf{Validation}, forming a closed-loop task-solving framework (see Algorithm~\ref{alg:select_then_decompose}).

\begin{itemize}
    \item \textbf{Selection Module}: Utilizes an LLM guided by a prompt \( P \) to analyze the input question \( Q \), selecting a suitable decomposition approach \( A \) along with reasoning \( R \).
    
    \item \textbf{Execution Module}: Applies the selected approach \( A \) to the input task \( Q \), producing a candidate solution \( S \).
    
    \item \textbf{Validation Module}: Assesses the confidence score \( C \in [0,1] \) of the solution \( S \) based on the original question \( Q \). If \( C \geq T \), the solution is accepted. Otherwise, the system initiates a staged switching mechanism, sequentially exploring \{IO\}, implicit approaches \{\text{CoT}, \text{P\&S}\}, and explicit approaches \{\text{ReAct}, \text{P\&E}, \text{P\&E (DAG)}\}, with uniform random sampling within each group.
\end{itemize}

This modular design enables flexible and efficient task resolution with adaptive cost-performance tradeoffs.
\begin{algorithm*}
\caption{Select-Then-Decompose Strategy}
\label{alg:select_then_decompose}
\begin{algorithmic}[1]
\Require Question \( Q \), Threshold \( T \), Max attempts \( K \), Instruction prompt \( P \), Define groups \( DG \): \{IO\}, Implicit \{\text{CoT}, \text{P\&S}\}, Explicit \{\text{ReAct}, \text{P\&E}, \text{P\&E (DAG)}\}

\Ensure Solution $S$ or $\emptyset$
\State $k \gets 0$, $S \gets \emptyset$
\While{$k < K$}
    \If{$k = 0$}
        \State $(A, R)\gets$\textcolor{red}{$\text{LLM}(Q,P_{seletion})$}\Comment{Generate selected method and reason}
    \Else
        \State $G \gets$ next group in $DG$\Comment{Staged order:IO$\rightarrow$Implicit$\rightarrow$Explicit}
        \State Sample $M$ randomly from $U(G)$
    \EndIf
    \State $S \gets$ \textcolor{blue}{$\text{Execute}(A, Q)$}
    \State $C \gets$ \textcolor{orange}{$\text{LLM}(Q, S,P_{validation})$}\Comment{Generate confidence score $C\in [0, 1]$}
    \If{$C \geq T$}
        \State \Return $S$\Comment{Generate a satisfactory solution}
    \Else
        \State $k \gets k + 1$\Comment{Update the attempt count}
    \EndIf
\EndWhile
\State \Return $\emptyset$\Comment{No appropriate solution}
\end{algorithmic}
\end{algorithm*}

\section{Complete Prompts}
\subsection{Prompt for Select-Then-Decompose Strategy}
\label{appendix: Prompt for Select-Then-Decompose Strategy}
For the Selection Module, to improve the LLM's decision-making ability, the prompt systematically introduces various decomposition approaches and guides the model to consider task complexity and semantic features when making a choice. 
We standardize the model's output format, requiring it to return two key elements: the name of the selected method $M$, and the reason for selection $R$.

For the Validation Module, to let LLM score the solutions purely without other interference, we simply prompt the big model to generate a confidence score between 0 and 1 based on the problem and solution, and also specify that it format the output Reason and score.
\begin{promptbox}[Prompt for Selection Module]{darkblue}
Please analyze the characteristics of the task description and select the most suitable method to solve the task from the candidate methods.\\

Task description: \{problem\}\\

Please analyze the characteristics of the task from the following dimensions:\\
- Whether it has clear goals and solution steps (logic).\\
- Whether it may require multiple rounds of attempts, corrections, or dynamic adjustments (iterative).\\
- Whether it involves information collection, viewpoint exploration (divergent).\\

Candidate methods and introduction:\\
- io: Input-Output, directly outputs the answer, suitable for simple problems.\\
- cot: Chain of Thought, step-by-step thinking and reasoning to generate answers, suitable for problems that require logical deduction.\\
- ps: Plan \& Solve, make a plan first and then execute, suitable for problems that require logical deduction.\\
- react: Reason+Act, alternate reasoning and execution, suitable for iterative tasks.\\
- pe: Plan \& Execute, generate a plan and execute it in sequence, suitable for vertical tasks with strict logic.\\
- dag\_flow: build a task structure of a directed acyclic graph, suitable for divergent tasks of parallel processing and extensive information collection.\\

When choosing a method, please combine the specific characteristics of the task with the applicable scenarios of the above methods to explain your reasons for choosing.\\

Please strictly follow the following format:\\
<think>\\
Your analysis\\
</think>\\
<answer>\\
Your choice (only fill in the method name, such as: cot, ps, etc.)\\
</answer>\\
\end{promptbox}

\begin{promptbox}[Prompt for Validation Module]{darkblue}
Please, as a serious evaluator, rate the quality of the following "solution".\\

Problem:\\
\{problem\}\\

Solution:\\
\{solution\}\\

Please give your **confidence score** for the solution, give your explanation, and return a floating point number between 0 and 1.\\

Please strictly follow the following format:\\
<think>\\
Your analysis\\
</think>\\
<score>\\
Your confidence score\\
</score>\\
\end{promptbox}

\subsection{Prompt for IO}
\begin{promptbox}[Prompt for IO]{darkblue}
    Q: \{question\}\\
    
    A: Please output the final answer directly.
\end{promptbox}

\subsection{Prompt for CoT}
\begin{promptbox}[Prompt for CoT (Zero Shot)]{darkblue}
    Q: \{question\}\\
    
    A: Let's think step by step.
\end{promptbox}

\subsection{Prompt for Plan-and-Solve}
\begin{promptbox}[Prompt for Plan-and-Solve]{darkblue}
    Q: \{question\}\\
    
    A: Let's first understand the problem and devise a plan to solve the problem. \\
        <Plan>\\
        Step 1. xxx\\
        Step 2. xxx\\
        ...(repeat as needed)\\
        </Plan>\\
        
    Then, let's carry out the plan to solve the problem step by step.\\
        <Solution>\\
        Place your solution for each step in the plan.\\
        </Solution>\\
\end{promptbox}

\subsection{Prompt for ReAcT}
\begin{promptbox}[Prompt for ReAcT]{darkblue}
    Answer the following questions as best you can. \\

    Use the following format:\\

    Question: the input question you must answer.\\
    Thought: you should always think about what to do.\\
    Subtask: your subtask to carry out.\\
    Result: the result of the subtask.\\
    ... (this Thought/Subtask/Result can repeat N times).\\
    Thought: I now know the final answer.\\
    Final Answer: the final answer to the original input question.\\

    Begin!\\

    Question: \{question\}\\
\end{promptbox}

\subsection{Prompt for Plan-and-Execute}
\begin{promptbox}[Prompt for Plan]{darkblue}
    Let's first understand the following problem and devise a linear plan to solve the problem.\\
    \{question\}\\

    Use the following format:\\
    Subtask 1: [First step to solve the problem]\\
    Subtask 2: [Second step to solve the problem]\\
    ...(repeat as needed)\\
    
    Provide only the subtasks as a plan. Do not execute or generate results for any subtask.\\
    Begin!\\
\end{promptbox}

\begin{promptbox}[Prompt for Execute]{darkblue}
    \{question\}\\
    
    \{context\}\\
    
    \{subtask\_id\}: \{subtask\_description\}\\
    
    Please execute this \{subtask\_id\} and provide the result:
\end{promptbox}

\subsection{Prompt for Plan-and-Execute (DAG)}
\begin{promptbox}[Prompt for Plan]{darkblue}
    Let's first understand the following problem and devise a directed acyclic graph (DAG) of subtasks to solve the problem.\\

    \{question\}\\
    
    Use the following JSON format to break down the problem into subtasks:\\
    
    \{subtasks\_example\}\\
    
    Rules:\\
    1. Each subtask must have a unique ID (e.g., "Subtask 1").\\
    2. Each subtask must have a clear description of what needs to be done.\\
    3. If a subtask depends on other subtasks, list their IDs in the "dependencies" field.\\
    4. Ensure the subtasks form a directed acyclic graph (DAG) with no circular dependencies.\\
    5. Provide only the JSON output. Do not include any additional text.\\
    
    Begin!\\
\end{promptbox}

\begin{promptbox}[Subtasks\_example]{darkblue}
\{\\
    "subtasks": [\\
        \{\\
            "id": "Subtask 1",\\
            "description": "[First step to solve the problem]",\\
            "dependencies": []\\
        \},\\
        \{\\
            "id": "Subtask 2",\\
            "description": "[Second step to solve the problem]",\\
            "dependencies": ["Subtask 1"]\\
        \},\\
        \{\\
            "id": "Subtask 3",\\
            "description": "[Third step to solve the problem]",\\
            "dependencies": ["Subtask 1", "Subtask 2"]\\
        \}\\
    ]\\
\}\\
\end{promptbox}

\begin{promptbox}[Prompt for Execute]{darkblue}
   Here is the original question: \{question\}\\

    Here is the context of previous subtasks and their results:\\
    \{context\}\\
    
    The current subtask is: \{subtask\_id\}: \{subtask\_description\}\\
    
    Dependencies: \{subtask\_dependencies\}\\
    
    Please execute this subtask and provide the result. If the subtask depends on previous subtasks, use their results to complete the task.\\
\end{promptbox}

\section{Original Data}
\subsection{Original Data for Five Objective Benchmarks}
\label{appendix: Original Data for Five Objective Benchmarks}

\begin{figure*}[t]  
    \centering
    \includegraphics[width=\linewidth]{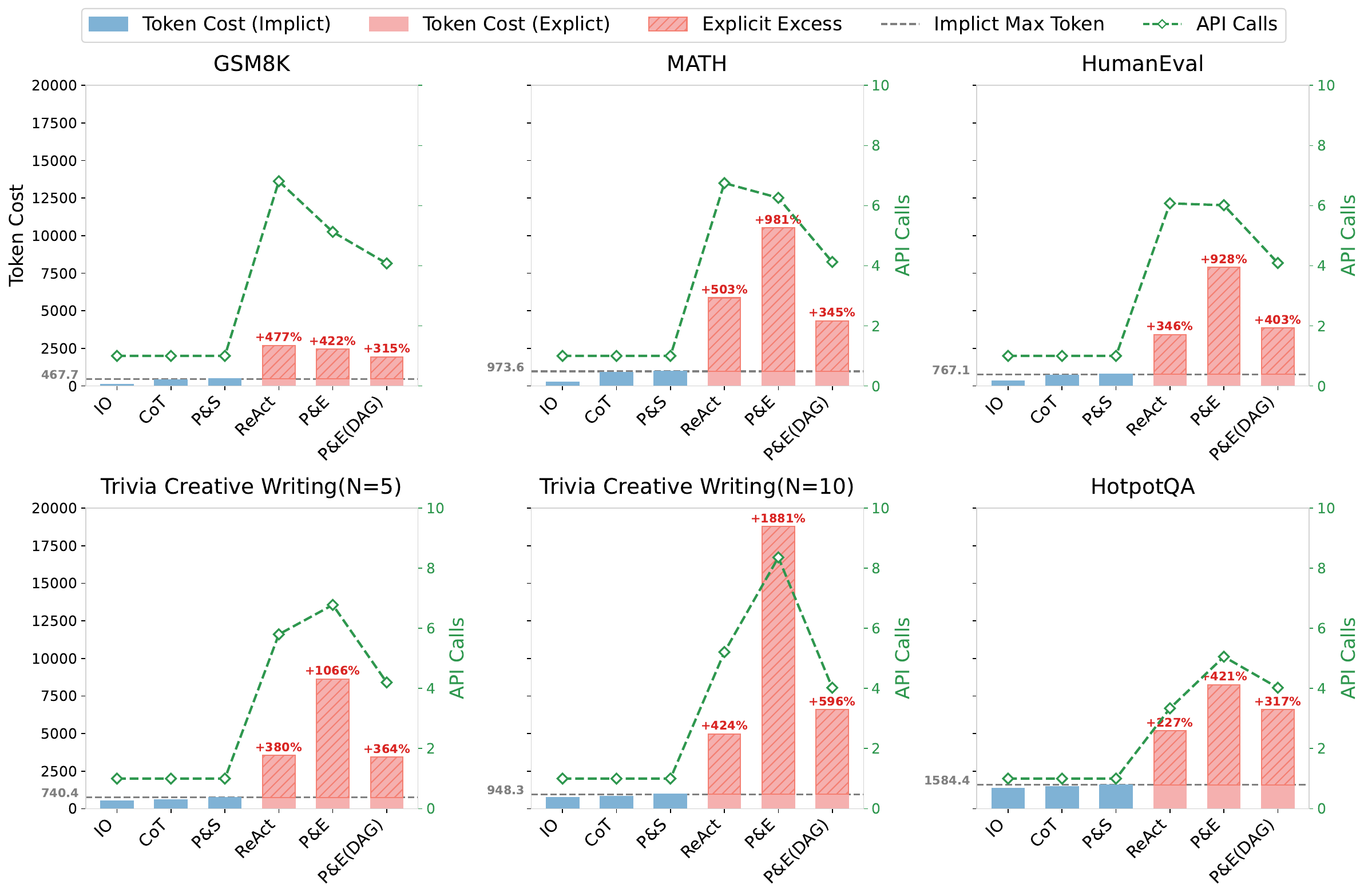}
    \caption{Token cost and API call analysis across all five benchmarks.}
    \label{fig:token_cost_2}
\end{figure*}

We present the raw data of various decomposition approaches on five benchmarks, including evaluation metrics, token consumption, and the number of API calls in Table~\ref{tab:Original Data for GSM8K}--\ref{tab:Original Data for HotpotQA}. Additionally, token cost and API call analysis across all five benchmarks are shown in Figure~\ref{fig:token_cost_2}. 

\begin{table}[H]
\centering
\caption{Original Data for GSM8K}
\label{tab:Original Data for GSM8K}
\resizebox{\columnwidth}{!}{
\begin{tabular}{lccc}
\toprule
Method & Accuracy (\%) & Tokens & API Calls \\
\midrule
IO & 33.13 & 106.37 & 1.00 \\
CoT  & 93.17 & 396.45 & 1.00 \\
P\&S  & 92.72 & 467.70 & 1.00 \\
ReAct  & 91.66 & 2700.04 & 6.81 \\
P\&E  & 92.34 & 2442.82 & 5.13 \\
P\&E (DAG)  & 90.67 & 1943.24 & 4.08 \\
\midrule
Select-Then-Decompose & 93.56 & 516.57 & 1.28 \\
\bottomrule
\end{tabular}
}
\end{table}

\begin{table}[H]
\centering
\caption{Original Data for MATH}
\resizebox{\columnwidth}{!}{
\begin{tabular}{lccc}
\toprule
Method & Accuracy (\%) & Tokens & API Calls \\
\midrule
IO & 17.02 & 240.25 & 1.00 \\
CoT & 50.73 & 890.47 & 1.00 \\
P\&S  & 49.10 & 973.55 & 1.00 \\
ReAct & 44.83 & 5865.86 & 6.75 \\
P\&E & 52.09 & 10521.10 & 6.26 \\
P\&E (DAG) & 48.06 & 4330.35 & 4.13 \\
\midrule
Select-Then-Decompose & 52.39 & 2560.22 & 2.31 \\
\bottomrule
\end{tabular}
}
\end{table}

\begin{table}[H]
\centering
\caption{Original Data for HumanEval}
\resizebox{\columnwidth}{!}{
\begin{tabular}{lccc}
\toprule
Method & Accuracy (\%) & Tokens & API Calls \\
\midrule
IO & 84.73 & 328.03 & 1.00 \\
CoT & 86.26 & 724.98 & 1.00 \\
P\&S & 84.73 & 767.08 & 1.00 \\
ReAct & 90.07 & 3421.78 & 6.08 \\
P\&E & 83.21 & 7887.10 & 6.02 \\
P\&E (DAG) & 83.97 & 3856.52 & 4.09 \\
\midrule
Select-Then-Decompose & 88.55 & 845.82 & 1.18 \\
\bottomrule
\end{tabular}
}
\end{table}

\begin{table}[H]
\centering
\caption{Original Data for Creative Writing(N=5)}
\resizebox{\columnwidth}{!}{
\begin{tabular}{lccc}
\toprule
Method & Score & Tokens & API Calls \\
\midrule
IO & 47.00 & 497.24 & 1.00 \\
CoT  & 49.80 & 591.00 & 1.00 \\
P\&S & 49.20 & 740.42 & 1.00 \\
ReAct  & 62.00 & 3552.34 & 5.80 \\
P\&E & 62.80 & 8630.23 & 6.78 \\
P\&E (DAG) & 64.60 & 3432.22 & 4.20 \\
\midrule
Select-Then-Decompose & 59.20 & 2902.72 & 1.39 \\
\bottomrule
\end{tabular}
}
\end{table}

\begin{table}[H]
\centering
\caption{Original Data for Creative Writing(N=10)}
\resizebox{\columnwidth}{!}{
\begin{tabular}{lccc}
\toprule
Method & Score & Tokens & API Calls \\
\midrule
IO & 51.50 & 720.74 & 1.00 \\
CoT & 51.20 & 815.77 & 1.00 \\
P\&S & 51.20 & 948.31 & 1.00 \\
ReAct & 63.00 & 4965.75 & 5.21 \\
P\&E & 54.20 & 18783.87 & 8.36 \\
P\&E (DAG) & 64.10 & 6596.20 & 4.02 \\
\midrule
Select-Then-Decompose & 55.70 & 1461.76 & 1.57 \\
\bottomrule
\end{tabular}
}
\end{table}

\begin{table}[H]
\centering
\caption{Original Data for HotpotQA}
\label{tab:Original Data for HotpotQA}
\resizebox{\columnwidth}{!}{
\begin{tabular}{lccc}
\toprule
Method & Score & Tokens & API Calls \\
\midrule
IO & 60.54 & 1348.62 & 1.00 \\
CoT & 63.00 & 1466.95 & 1.00 \\
P\&S & 61.93 & 1584.43 & 1.00 \\
ReAct & 53.52 & 5181.09 & 3.34 \\
P\&E & 63.22 & 8254.73 & 5.06 \\
P\&E (DAG) & 65.52 & 6600.19 & 4.02 \\
\midrule
Select-Then-Decompose & 65.26 & 1639.99 & 1.14 \\
\bottomrule
\end{tabular}
}
\end{table}

\subsection{Original Data for MT-Bench}
\label{appendix: Original Data for MT-Bench}

We present the raw data of five classic decomposition approaches on MT-Bench in Table~\ref{tab:Original Data for MT-Bench: Writing}--\ref{tab:Original Data for MT-Bench: Coding}, including scores of two turns, token consumption, and the number of API calls.

\begin{table}[H]
\centering
\caption{Original Data for MT-Bench: Writing}
\label{tab:Original Data for MT-Bench: Writing}
\resizebox{\columnwidth}{!}{
\begin{tabular}{lcccc}
\toprule
Method & Turn 1 & Turn 2 & Tokens & API Calls \\
\midrule
IO & 8.92 & 7.63 & 1031.20 & 2.00 \\
CoT & 9.06 & 7.38 & 1156.77 & 2.00 \\
P\&S & 8.91 & 8.03 & 1292.13 & 2.00 \\
ReAct & 8.10 & 7.98 & 7170.17 & 9.93 \\
P\&E & 9.05 & 7.04 & 22482.47 & 14.17 \\
P\&E (DAG) & 9.27 & 7.78 & 7635.53 & 8.07 \\

\bottomrule
\end{tabular}
}
\end{table}

\begin{table}[H]
\centering
\caption{Original Data for MT-Bench: Roleplay}
\resizebox{\columnwidth}{!}{
\begin{tabular}{lcccc}
\toprule
Method & Turn 1 & Turn 2 & Tokens & API Calls \\
\midrule
IO & 8.28 & 6.96 & 803.70 & 2.00 \\
CoT & 8.50 & 6.98 & 1312.97 & 2.00 \\
P\&S & 8.90 & 7.06 & 1616.60 & 2.00 \\
ReAct & 6.77 & 6.23 & 5195.17 & 9.17 \\
P\&E & 8.53 & 6.39 & 36750.83 & 14.67 \\
P\&E (DAG) & 8.80 & 7.77 & 10553.40 & 8.00 \\

\bottomrule
\end{tabular}
}
\end{table}

\begin{table}[H]
\centering
\caption{Original Data for MT-Bench: Reasoning}
\resizebox{\columnwidth}{!}{
\begin{tabular}{lcccc}
\toprule
Method & Turn 1 & Turn 2 & Tokens & API Calls \\
\midrule
IO & 5.85 & 6.05 & 308.03 & 2.00 \\
CoT & 8.55 & 7.66 & 1037.83 & 2.00 \\
P\&S & 7.61 & 7.00 & 1200.50 & 2.00 \\
ReAct & 8.08 & 7.12 & 3544.23 & 6.93 \\
P\&E & 8.13 & 6.02 & 17796.23 & 11.33 \\
P\&E (DAG) & 8.08 & 6.83 & 9270.87 & 8.10 \\

\bottomrule
\end{tabular}
}
\end{table}

\begin{table}[H]
\centering
\caption{Original Data for MT-Bench: Math}
\resizebox{\columnwidth}{!}{
\begin{tabular}{lcccc}
\toprule
Method & Turn 1 & Turn 2 & Tokens & API Calls \\
\midrule
IO & 4.75 & 3.87 & 239.70 & 2.00 \\
CoT & 9.73 & 8.55 & 1416.63 & 2.00 \\
P\&S & 9.63 & 7.93 & 1615.77 & 2.00 \\
ReAct & 9.58 & 7.32 & 7822.17 & 11.50 \\
P\&E & 9.33 & 7.70 & 15064.93 & 11.13 \\
P\&E (DAG) & 9.47 & 7.50 & 9702.77 & 8.27 \\

\bottomrule
\end{tabular}
}
\end{table}

\begin{table}[H]
\centering
\caption{Original Data for MT-Bench: Coding}
\label{tab:Original Data for MT-Bench: Coding}
\resizebox{\columnwidth}{!}{
\begin{tabular}{lcccc}
\toprule
Method & Turn 1 & Turn 2 & Tokens & API Calls \\
\midrule
IO & 5.15 & 4.50 & 921.80 & 2.00 \\
CoT & 7.82 & 5.69 & 1762.10 & 2.00 \\
P\&S & 6.25 & 5.90 & 1844.20 & 2.00 \\
ReAct & 6.00 & 6.70 & 5231.80 & 9.20 \\
P\&E & 4.95 & 5.27 & 24217.20 & 12.90 \\
P\&E (DAG) & 4.86 & 4.87 & 8388.40 & 8.20 \\

\bottomrule
\end{tabular}
}
\end{table}

\subsection{Original Data for DROP}
\label{appendix: Original Data for DROP}

\begin{table}[htbp]
\centering
\caption{Original Data for DROP}
\label{tab:drop}
\resizebox{\columnwidth}{!}{
\begin{tabular}{lcccc}
\toprule
\textbf{Metric} & \textbf{IO} & \textbf{CoT} & \textbf{P\&S} & \textbf{ReAct} \\
\midrule
Performance(\%) & 66.80 & 72.40 & 71.66 & 70.49 \\
Token Cost  & 319.31 & 471.16 & 567.41 & 2017.96 \\
\bottomrule
\end{tabular}
}

\vspace{0.3em} 
\resizebox{\columnwidth}{!}{
\begin{tabular}{lccc}
\toprule
\textbf{Metric} & \textbf{P\&E} & \textbf{P\&E (DAG)} & \textbf{S\&D} \\
\midrule
Performance(\%) & 75.30 & 76.58 & 74.34 \\
Token Cost  & 2424.74 & 2461.81 & 851.35 \\
\bottomrule
\end{tabular}
}
\end{table}

\newpage
\section{Examples}
\label{appendix: examples}
\begin{promptbox}[Example of IO in GSM8K]{lightgreen}

\textbf{Question:}\\

Janet sells \( 16 - 3 - 4 = 9 \) duck eggs a day. She makes \( 9 \times 2 = 18 \) dollars every day at the farmer's market.

\bigskip

\textbf{Answer:}\\

Janet makes \$18 every day at the farmer's market.
    
\end{promptbox}

\begin{promptbox}[Example of CoT in MATH]{lightgreen}
    \textbf{Question:}\\
    
    What is the value of \( x \) if \( -\frac{2}{3}(x-5) = \frac{3}{2}(x+1) \)?

\bigskip

\textbf{Answer:}\\

To solve the equation \( -\frac{2}{3}(x-5) = \frac{3}{2}(x+1) \), we will follow these steps:

\begin{enumerate}
  \item \textbf{Distribute both sides}:
    \begin{align*}
    -\frac{2}{3}(x-5) &= -\frac{2}{3}x + \frac{10}{3} \\
    \frac{3}{2}(x+1) &= \frac{3}{2}x + \frac{3}{2}
    \end{align*}
    
    So the equation becomes:
    \[
    -\frac{2}{3}x + \frac{10}{3} = \frac{3}{2}x + \frac{3}{2}
    \]

  \item \textbf{Eliminate the fractions}:

  Multiply both sides of the equation by the least common multiple (LCM) of the denominators, which is 6:
  \[
  6\left(-\frac{2}{3}x + \frac{10}{3}\right) = 6\left(\frac{3}{2}x + \frac{3}{2}\right)
  \]

  This simplifies to:
  \[
  -4x + 20 = 9x + 9
  \]

  \item \textbf{Combine like terms}:

  Move all terms involving \( x \) to one side and constants to the other:
  \begin{align*}
  -4x - 9x &= 9 - 20 \\
  -13x &= -11
  \end{align*}

  \item \textbf{Solve for \( x \)}:

  Divide both sides by \(-13\):
  \[
  x = \frac{-11}{-13} = \frac{11}{13}
  \]
\end{enumerate}

\bigskip

Thus, the value of \( x \) is \( \boxed{\frac{11}{13}} \).
\end{promptbox}

\begin{promptbox}[Example of P\&S in HotpotQA]{lightgreen}
    \textbf{Question:} \\
    
    "Constantin Medien AG (formerly \textit{EM.Entertainment} and \href{http://em.tv/}{EM.TV \& Merchandising AG}, then \href{http://em.tv/}{EM.TV AG}, and finally \textit{em.sport media ag}) is a German media group, based in Ismaning near Munich, active in the area of sports, film, and event marketing to medium-sized media companies."

    ...(Omit the subsequent text)\\

Read and understand the above text, and answer the following questions.\\

IVA Media AG changed its name in 2004. What does their new acronym stand for?

\bigskip

\textbf{Answer:}\\

\textbf{Plan}
\begin{itemize}
    \item Step 1: Identify the new name of VIVA Media AG after it changed its name in 2004.
    \item Step 2: Determine if the new name has a specific acronym that needs to be explained.
    \item Step 3: Provide the full form of the acronym if it stands for something specific.
\end{itemize}

\bigskip

\textbf{Solution}
\begin{itemize}
    \item Step 1: The name of VIVA Media AG was changed to VIVA Media GmbH in 2004.
    \item Step 2: The acronym \texttt{"GmbH"} stands for \textit{"Gesellschaft mit beschränkter Haftung,"} which is a type of legal entity in Germany and other German-speaking countries that denotes a company with limited liability.
    \item Step 3: Therefore, the new acronym \texttt{"GmbH"} specifically indicates the legal structure of the company rather than an acronym with a unique meaning related to the media industry.
\end{itemize}

\bigskip

\textbf{Result} \\
Gesellschaft mit beschränkter Haftung
\end{promptbox}

\begin{promptbox}[Example of ReAcT in HumanEval]{lightgreen}
    \textbf{Question:}

\begin{lstlisting}
def solve(N):
"""Given a positive integer N, return the total sum of its digits in binary.

Example
    For N = 1000, the sum of digits will be 1 the output should be "1".
    For N = 150, the sum of digits will be 6 the output should be "110".
    For N = 147, the sum of digits will be 12 the output should be "1100".

Variables:
    @N integer
    sConstraints: 0 <= N <=le 10000.
Output:
     a string of binary number
"""
\end{lstlisting}

\textbf{Answer:}\\

\textbf{Thought:} I need to implement a function that calculates the sum of the digits of a given integer \( N \) and then converts that sum into its binary representation.

\textbf{Subtask:} First, I will calculate the sum of the digits of \( N \).

\textbf{Result:} To calculate the sum of the digits of \( N \), I will convert \( N \) to a string, iterate through each character, convert it back to an integer, and sum these integers.

\begin{lstlisting}
def sum_of_digits(N):
    return sum(int(digit) for digit in str(N))
\end{lstlisting}

Now, I will apply this function to \( N \).

\textbf{Thought:} I need to apply the \texttt{sum\_of\_digits} function to the input \( N \) to get the total sum of its digits.

\textbf{Subtask:} I will call the \texttt{sum\_of\_digits} function with \( N \) as the argument.

\textbf{Result:}
\begin{lstlisting}
digit_sum = sum_of_digits(N)
\end{lstlisting}

\textbf{Thought:} I now need to convert the calculated digit sum into its binary representation.

\textbf{Subtask:} I will use Python's built-in \texttt{bin} function, which gives the binary representation of an integer. The result includes a \texttt{'0b'} prefix, so I'll slice off the first two characters.

\textbf{Result:}
\begin{lstlisting}
binary_representation=bin(digit_sum)[2:]
\end{lstlisting}

\textbf{Final Thought:} I now know the final answer.

\textbf{Final Answer:} The final answer to the original input question is the implementation of the function \texttt{solve(N)}:

\begin{lstlisting}
def solve(N):
    def sum_of_digits(N):
        return sum(int(digit) for digit in str(N))
    digit_sum = sum_of_digits(N)
    binary_representation = bin(digit_sum)[2:]
    return binary_representation
\end{lstlisting}
\end{promptbox}

\begin{promptbox}[Example of P\&E in MATH]{lightgreen}

\textbf{Question:}\\

Find the smallest positive integer that is greater than \(1\) and relatively prime to the product of the first 20 positive integers.  
Reminder: two numbers are relatively prime if their greatest common divisor is 1.

\vspace{1em}

\textbf{Answer:}\\

\textbf{Plan:}

\begin{itemize}
  \item Subtask 1: Calculate the product of the first 20 positive integers, known as \(20!\).
  \item Subtask 2: Determine the prime factorization of \(20!\).
  \item Subtask 3: Identify the smallest positive integer greater than 1 that is not included in the prime factorization of \(20!\).
  \item Subtask 4: Verify that the identified integer is relatively prime to \(20!\) by checking its greatest common divisor with \(20!\).
  \item Subtask 5: Conclude and report the smallest positive integer greater than 1 that is relatively prime to the product of the first 20 positive integers.
\end{itemize}

\vspace{1em}

\textbf{Execution:}

\textbf{Subtask 1: Calculate \(20!\).}

\[
20! = 20 \times 19 \times 18 \times \cdots \times 2 \times 1
\]

Calculating stepwise:

\[
\begin{aligned}
1 \times 2 &= 2 \\
2 \times 3 &= 6 \\
6 \times 4 &= 24 \\
24 \times 5 &= 120 \\
120 \times 6 &= 720 \\
720 \times 7 &= 5040 \\
5040 \times 8 &= 40320 \\
40320 \times 9 &= 362880 \\
362880 \times 10 &= 3628800 \\
3628800 \times 11 &= 39916800 \\
39916800 \times 12 &= 479001600 \\
479001600 \times 13 &= 6227020800 \\
6227020800 \times 14 &= 87178291200 \\
87178291200 \times 15 &= 1307674368000 \\
1307674368000 \times 16 &= 20922789888000 \\
20922789888000 \times 17 &= 355687428096000 \\
355687428096000 \times 18 &= 6402373705728000 \\
6402373705728000 \times 19 &= 121645100408832000 \\
121645100408832000 \times 20 &= 2432902008176640000
\end{aligned}
\]

Therefore,

\[
20! = 2432902008176640000
\]

\vspace{1em}

\end{promptbox}

\begin{promptbox}[Example of P\&E in MATH]{lightgreen}

\textbf{Subtask 2: Prime factorization of \(20!\).}

The prime numbers up to 20 are: \(2, 3, 5, 7, 11, 13, 17, 19\).

The exponent of a prime \(p\) in \(n!\) is given by

\[
\sum_{k=1}^{\infty} \left\lfloor \frac{n}{p^k} \right\rfloor.
\]

Calculate each exponent for \(20!\):

\[
\begin{aligned}
\text{For } p=2: & \quad \left\lfloor \frac{20}{2} \right\rfloor + \left\lfloor \frac{20}{4} \right\rfloor + \left\lfloor \frac{20}{8} \right\rfloor + \left\lfloor \frac{20}{16} \right\rfloor \\
& = 10 + 5 + 2 + 1 = 18 \\
\text{For } p=3: & \quad \left\lfloor \frac{20}{3} \right\rfloor + \left\lfloor \frac{20}{9} \right\rfloor = 6 + 2 = 8 \\
\text{For } p=5: & \quad \left\lfloor \frac{20}{5} \right\rfloor + \left\lfloor \frac{20}{25} \right\rfloor = 4 + 0 = 4 \\
\text{For } p=7: & \quad \left\lfloor \frac{20}{7} \right\rfloor = 2 \\
\text{For } p=11: & \quad \left\lfloor \frac{20}{11} \right\rfloor = 1 \\
\text{For } p=13: & \quad \left\lfloor \frac{20}{13} \right\rfloor = 1 \\
\text{For } p=17: & \quad \left\lfloor \frac{20}{17} \right\rfloor = 1 \\
\text{For } p=19: & \quad \left\lfloor \frac{20}{19} \right\rfloor = 1 \\
\end{aligned}
\]

Hence,

\[
20! = 2^{18} \times 3^{8} \times 5^{4} \times 7^{2} \times 11 \times 13 \times 17 \times 19.
\]

\vspace{1em}

\textbf{Subtask 3: Identify the smallest positive integer greater than 1 relatively prime to \(20!\).}

Since \(20!\) includes all prime factors up to 19, the smallest integer greater than 1 that shares no prime factors with \(20!\) must be a prime greater than 19.

The smallest such prime is

\[
\boxed{23}.
\]

\vspace{1em}

\textbf{Subtask 4: Verify \(\gcd(23, 20!) = 1\).}

Because \(23\) is prime and greater than all prime factors of \(20!\), it shares no common prime factor with \(20!\).

Therefore,

\[
\gcd(23, 20!) = 1,
\]

which means \(23\) is relatively prime to \(20!\).

\vspace{1em}

\textbf{Subtask 5: Conclusion}

The smallest positive integer greater than 1 that is relatively prime to the product of the first 20 positive integers is

\[
\boxed{23}.
\]

\end{promptbox}

\begin{promptbox}[Example of P\&E (DAG) in Trivia Creative Writing]{lightgreen}
\textbf{Question:}\\

Write a short and coherent story about Harry Potter that incorporates the answers to the following 5 questions: Who was the man behind The Chipmunks? Which Lloyd Webber musical premiered in the US on 10th December 1993? Who was the next British Prime Minister after Arthur Balfour? Who had a 70s No. 1 hit with Kiss You All Over? What claimed the life of singer Kathleen Ferrier?\\

\textbf{Answer:}\\

\textbf{Plan:}

\begin{lstlisting}     
{
    "subtasks": [
        {
            "id": "Subtask 1",
            "description": "Identify the five key questions that must be answered to create a story about Harry Potter.",
            "dependencies": []
        },
        {
            "id": "Subtask 2",
            "description": "Research the answers to each of the five questions: The man behind The Chipmunks, the Lloyd Webber musical, the next British Prime Minister after Arthur Balfour, the artist of 'Kiss You All Over', and the cause of Kathleen Ferrier's death.",
            "dependencies": ["Subtask 1"]
        },
        {
            "id": "Subtask 3",
            "description": "Compose a coherent story about Harry Potter that integrates the answers to the five questions found in Subtask 2.",
            "dependencies": ["Subtask 1", "Subtask 2"]
        }
    ]
}
    \end{lstlisting}

\textbf{Execution:}

\textbf{Subtask 1: Identify the five key questions that must be answered to create a story about Harry Potter. Dependencies: []}

\textbf{Result:} To create a story about Harry Potter that incorporates the answers to the specified questions, the five key questions to be answered are:

\end{promptbox}

\begin{promptbox}[Example of P\&E (DAG) in Trivia Creative Writing]{lightgreen}
    
\begin{enumerate}
    \item Who was the man behind The Chipmunks?
    \item Which Lloyd Webber musical premiered in the US on 10th December 1993?
    \item Who was the next British Prime Minister after Arthur Balfour?
    \item Who had a 70s No 1 hit with ``Kiss You All Over''?
    \item What claimed the life of singer Kathleen Ferrier?
\end{enumerate}

These questions provide the necessary details that need to be woven into the narrative about Harry Potter.

\textbf{Subtask 2: Research the answers to each of the five questions. Dependencies: [Subtask 1]}

\textbf{Result:} Here are the answers to the five key questions:

\begin{enumerate}
    \item \textbf{Who was the man behind The Chipmunks?}\\
    The man behind The Chipmunks was Ross Bagdasarian Sr., who created the characters and the music that made them famous.
    
    \item \textbf{Which Lloyd Webber musical premiered in the US on 10th December 1993?}\\
    The Lloyd Webber musical that premiered in the US on 10th December 1993 is \textit{Sunset Boulevard}.
    
    \item \textbf{Who was the next British Prime Minister after Arthur Balfour?}\\
    The next British Prime Minister after Arthur Balfour was David Lloyd George.
    
    \item \textbf{Who had a 70s No 1 hit with ``Kiss You All Over''?}\\
    The artist who had a 70s No. 1 hit with ``Kiss You All Over'' is Exile.
    
    \item \textbf{What claimed the life of singer Kathleen Ferrier?}\\
    Kathleen Ferrier's life was claimed by cancer, specifically breast cancer.
\end{enumerate}

\textbf{Subtask 3: Compose a coherent story about Harry Potter that integrates the answers to the five questions. Dependencies: [Subtask 1, Subtask 2]}

\textbf{Result:}

Once upon a time in the magical world of Hogwarts, Harry Potter found himself in a peculiar situation. It all began when he stumbled upon an old, dusty book in the library titled \textit{The Enchanted History of Muggle Music.} Intrigued, Harry opened it to find a chapter dedicated to the legendary Ross Bagdasarian Sr., the man behind The Chipmunks. The book described how Bagdasarian had enchanted the world with his high-pitched melodies, and Harry couldn't help but imagine what it would be like to hear the Chipmunks sing a wizarding song.

As he flipped through the pages, Harry discovered a mention of a famous musical that had premiered in the Muggle world on December 10, 1993—Andrew Lloyd Webber's \textit{Sunset Boulevard.} The story of a faded star and her struggles resonated with Harry, who often felt the weight of expectations on his shoulders. He wondered if the characters in the musical ever faced dark wizards or had to navigate the complexities of friendship like he did.

...(Omit the subsequent story)

\end{promptbox}

\begin{promptbox}[Example of S\&D (No Switch)]{lightgreen}

\textbf{Question:}\\

For Halloween, Megan received 11 pieces of candy from neighbors and 5 pieces from her older sister. If she only ate 8 pieces a day, how long would the candy last her?\\

\textbf{Answer:}\\

\textbf{Selection Module:}\\

<think>

The task has a clear goal: to determine how long Megan's candy will last given the total amount of candy and her daily consumption rate. The solution steps involve straightforward arithmetic: first, calculating the total amount of candy and then dividing that by the daily consumption. This indicates a logical deduction process, which aligns well with methods that emphasize step-by-step reasoning.

The task does not require multiple rounds of attempts or dynamic adjustments, as it is a straightforward calculation without the need for iterative refinement. Therefore, it does not fit the characteristics of an iterative method.

Additionally, the task does not involve extensive information collection or viewpoint exploration, as it is focused on a single calculation rather than exploring multiple perspectives or gathering varied data.

Given these characteristics, the most suitable method for this task is one that emphasizes logical deduction and step-by-step reasoning, making the Chain of Thought (cot) method the best fit.

</think>\\

<answer>\\
\textbf{cot}\\
</answer>\\

\textbf{Execution Module Process (CoT):}

\begin{enumerate}
    \item \textbf{Total Candy Calculation}:
    \begin{itemize}
        \item Megan received 11 pieces of candy from neighbors.
        \item She also received 5 pieces from her older sister.
        \item Total candy:
        \[
        11 + 5 = 16 \text{ pieces of candy}
        \]
    \end{itemize}
    
    \item \textbf{Daily Consumption}:
    \begin{itemize}
        \item Megan eats 8 pieces of candy per day.
    \end{itemize}
    
    \item \textbf{Duration Calculation}:
    \begin{itemize}
        \item To find out how many days the candy will last, divide the total amount of candy by the amount she eats per day:
        \[
        \frac{16 \text{ pieces}}{8 \text{ pieces/day}} = 2 \text{ days}
        \]
    \end{itemize}
\end{enumerate}

\textbf{Validation Module:}

\end{promptbox}

\begin{promptbox}[Example of S\&D (No Switch) ]{lightgreen}

<think>

The solution provided is accurate and follows a logical step-by-step approach to solve the problem. It correctly calculates the total amount of candy Megan has by adding the pieces received from neighbors and her sister. The daily consumption is also correctly stated, and the final calculation to determine how many days the candy will last is performed correctly by dividing the total candy by the daily consumption. The conclusion that Megan's candy will last her 2 days is correct based on the calculations provided. Therefore, I would rate the quality of this solution as very high.

</think>\\

<score>\\
1.0\\
</score>\\

Confidence Score: 1.0\\

\begin{lstlisting}
confidence_score=1.0 > threshold=0.7
\end{lstlisting}

Validate: True, no switching required\\

\textbf{Final Approach: } CoT

\end{promptbox}

\begin{promptbox}[Example of S\&D (Switched)]{lightgreen}
\textbf{Question:}\\

Write a short and coherent story about Pikachu that incorporates the answers to the following 5 questions: Who directed the classic 30s western Stagecoach? Dave Gilmore and Roger Waters were in which rock group? Which highway was revisited in a classic '60s album by Bob Dylan? Which was the only Eastern Bloc country to participate in the 1984 LA Olympics? Which 90s sci-fi series with James Belushi was based on Bruce Wagner's comic strip of the same name?\\

\textbf{Answer:}\\

\textbf{Selection Module:}\\

<think>\\
The task requires writing a coherent story about Pikachu while incorporating answers to five specific questions. This involves several characteristics:

1. **Clear Goals and Solution Steps**: The task has a clear goal of creating a story that includes specific information. However, the steps to achieve this are not strictly linear, as the story must weave in the answers creatively.

2. **Iterative Nature**: The task may require multiple rounds of attempts and adjustments to ensure that the story flows well and incorporates the answers seamlessly. This suggests an iterative approach might be beneficial.

3. **Information Collection and Viewpoint Exploration**: The task involves collecting specific information (the answers to the questions) and creatively exploring how to integrate that information into a narrative. This indicates a need for a method that can handle divergent thinking.

Given these characteristics, the most suitable method would be one that allows for both planning and iterative refinement, as well as the ability to explore different narrative angles. The "Plan \& Solve" method (PS) is appropriate here because it allows for an initial plan to be made regarding how to structure the story and then execute it while being open to adjustments as needed.

Thus, the method that best fits the task is "ps" as it combines planning with the flexibility to adapt the story as it is being written.\\
</think>\\

<answer>\\
ps\\
</answer>\\

\textbf{Execution Module Process (P\&S)}:\\

\textbf{Plan:}\\

Step 1: Introduce Pikachu in a setting that connects to the themes of adventure and exploration.\\
Step 2: Incorporate elements from the questions into the story, ensuring they flow naturally.\\
Step 3: Conclude the story with a resolution that ties back to Pikachu's journey.\\

\textbf{Solution:}\\

...(Omitted steps)\\

Once upon a time in the vibrant world of Pokémon, Pikachu found himself yearning for adventure beyond the familiar fields of Pallet Town. 

\end{promptbox}

\begin{promptbox}[Example of S\&D (Switched)]{lightgreen}

One sunny morning, he decided to embark on a journey along the famous Route 66, a highway that had been revisited in many tales, including a classic album by Bob Dylan. With a spark of excitement, Pikachu set off, his cheeks crackling with electricity. \\
... (Omit the subsequent story)\\





\textbf{Validation Module:}\\

<think>\\
The solution effectively incorporates the required elements from the questions into a coherent story about Pikachu. The references to Route 66 and Bob Dylan's album are smoothly integrated, as is the mention of the classic western "Stagecoach" directed by John Ford. The inclusion of Pink Floyd and the Eastern Bloc country participating in the 1984 LA Olympics adds depth to the narrative, while the reference to the sci-fi series "The Last Precinct" provides a fitting climax to Pikachu's adventure.

However, while the story is creative and engaging, some connections may feel a bit forced or less natural, particularly the transition between the various elements. The overall flow is good, but the integration of the trivia could be more seamless. Additionally, the story could benefit from more character development or emotional depth to enhance the reader's connection to Pikachu's journey.

Overall, the solution meets the requirements of the prompt and presents a fun and imaginative story, but there is room for improvement in the integration of the trivia elements and character depth.

</think>\\

<score>\\
0.65\\
</score>\\

Confidence Score: 0.65
\begin{lstlisting}
confidence_score=0.65 < threshold=0.7
\end{lstlisting}
Validate: False, switching to the next group of approaches... The next approach is P\&E.\\

\textbf{Execution Module Process (P\&E): }\\

\textbf{Plan:}\\

Subtask 1: Identify the key elements from each of the five questions that need to be incorporated into the story about Pikachu.

Subtask 2: Research and confirm the answers to the five questions to ensure accuracy.

Subtask 3: Create a rough outline of the story, determining how Pikachu will interact with the elements from the answers.   

Subtask 4: Write a draft of the story, weaving in the answers to the questions in a coherent and engaging manner.

Subtask 5: Revise the draft to improve flow, coherence, and ensure all elements are included appropriately.\\

\textbf{Execution: ... (Omit)}\\

**Pikachu's Adventure on Highway 61**

One sunny day in the Pokémon world, Pikachu was feeling particularly adventurous. 

\end{promptbox}

\begin{promptbox}[Example of S\&D (Switched)]{lightgreen}

He had heard tales of a legendary place called Highway 61, where trainers from all over gathered to challenge each other and share stories. Inspired by the grand adventures depicted in classic films, especially those directed by John Ford, who had once captured the spirit of the wild west in his iconic film "Stagecoach," Pikachu decided it was time to embark on his own journey. \\
... (Omit the subsequent story)\\

\textbf{Validation Module: }\\

<think>\\
The solution provided is a well-structured and coherent story that successfully incorporates the answers to the five questions into a narrative about Pikachu. The story flows logically, with each element being integrated in a way that enhances the overall adventure. The key elements from the questions are clearly represented: John Ford's influence is acknowledged, Pink Floyd's music is featured, Highway 61 is the setting, Romania's Olympic participation is mentioned, and the concept of "The Tenth Dimension" is creatively woven into the plot. The narrative captures the essence of adventure and friendship, which aligns well with Pikachu's character. Overall, the execution is strong, and the story is engaging and imaginative.
</think>\\

<score>\\
0.95\\
</score>\\

Confidence Score: 0.95
\begin{lstlisting}
confidence_score=0.95 > threshold=0.7
\end{lstlisting}
Validate: True,  no switching required\\

\textbf{Final Approach: } P\&E
\end{promptbox}